%% file: neurips_2022.tex
\documentclass{article}

\usepackage[final]{neurips_2022}

\usepackage[utf8]{inputenc} 
\usepackage[T1]{fontenc}    
\usepackage[pagebackref=true]{hyperref}
\usepackage{url}            
\usepackage{booktabs}       
\usepackage{amsfonts}       
\usepackage{nicefrac}       
\usepackage{microtype}      
\usepackage{xcolor}         
\usepackage{amsmath}
\usepackage{amssymb}
\usepackage{graphicx}
\usepackage{hyperref}
\usepackage[utf8]{inputenc}
\usepackage{natbib}
\usepackage{xcolor}
\usepackage{relsize}
\usepackage{ifthen}
\usepackage[normalem]{ulem}
\usepackage{svg}
\usepackage{subfiles}
\usepackage{xspace}
\usepackage{cleveref}
\usepackage{wrapfig}
\usepackage{dialogue}
\usepackage[subtle]{savetrees}
\usepackage{xurl}
\usepackage[most]{tcolorbox}
\definecolor{ForestGreen}{rgb}{0, 0.50, 0}

\input{std-macros}
\input{macros}
\renewcommand{\cite}{\citep}
\renewcommand*\backref[1]{\ifx#1\relax \else (Cited on #1) \fi}

\title{Picking on the Same Person: \\
Does Algorithmic Monoculture lead to Outcome Homogenization?}

\author{
Rishi Bommasani\thanks{Corresponding author.} \\
Computer Science\\
Stanford University\\
\texttt{nlprishi@stanford.edu} \\
\And 
Kathleen A. Creel \\
Philosophy, Computer Science\\
Northeastern University\\
\texttt{k.creel@northeastern.edu} \\
\And 
Ananya Kumar \\
Computer Science\\
Stanford University\\
\texttt{ananya@cs.stanford.edu} \\
\And 
Dan Jurafsky \\
Linguistics, Computer Science\\
Stanford University\\
\texttt{jurafsky@stanford.edu} \\
\And 
Percy Liang \\
Computer Science\\
Stanford University\\
\texttt{pliang@cs.stanford.edu} \\
}

\begin{document}
\maketitle

\input{sections/0-abstract}
\input{sections/1-introduction}
\input{sections/2-outcome-homogenization}
\input{sections/3-metrics}
\input{sections/4-data-sharing}
\input{sections/5-model-sharing}
\input{sections/6-societal-considerations}
\input{sections/7-conclusion}
\paragraph{Reproducibility.} All code, data, and experiments are available on GitHub and CodaLab Worksheets.\footnote{\url{https://worksheets.codalab.org/worksheets/0x807c29f8eb574d1fba8f429ec78b5d1b}}

\input{sections/12-acknowledgements}

\bibliographystyle{unsrtnat}
\bibliography{neurips}

\newpage
\input{sections/checklist}
\newpage
\appendix
\input{appendices/metrics}
\input{appendices/reproducibility}
\input{appendices/additional_experiments}

\end{document}

%% file: std-macros.tex
\DeclareMathOperator*{\var}{\text{Var}}
\newcommand{\cov}{\text{Cov}} 

\newcommand{\1}{\mathbb{I}} 

\ifthenelse{\isundefined{\definition}}{\newtheorem{definition}{Definition}}{}
\ifthenelse{\isundefined{\assumption}}{\newtheorem{assumption}{Assumption}}{}
\ifthenelse{\isundefined{\hypothesis}}{\newtheorem{hypothesis}{Hypothesis}}{}
\ifthenelse{\isundefined{\proposition}}{\newtheorem{proposition}{Proposition}}{}
\ifthenelse{\isundefined{\theorem}}{\newtheorem{theorem}{Theorem}}{}
\ifthenelse{\isundefined{\lemma}}{\newtheorem{lemma}{Lemma}}{}
\ifthenelse{\isundefined{\corollary}}{\newtheorem{corollary}{Corollary}}{}
\ifthenelse{\isundefined{\alg}}{\newtheorem{alg}{Algorithm}}{}
\ifthenelse{\isundefined{\example}}{\newtheorem{example}{Example}}{}
\newcommand{\E}{\mathop{\mathbb{E}}}

%% file: macros.tex
\newcommand\eg{e.g.~}
\newcommand\ie{i.e.~}

\newcommand\Havg{$H_{\textbf{avg}}$}
\newcommand\Hunif{$H_{\textbf{unif}}$}
\newcommand\Hworst{$H_{\textbf{worst}}$}
\newcommand\OH{$H^{\textbf{individual}}$}
\newcommand\OHG{$H_G^{\textbf{group}}$}

\newcommand\fail{\textsc{fail}}
\newcommand\systemic{\textsc{systemic failure}}

\def\Snospace~{\S{}}

\def\stable{0}

\if\stable1
\else
\fi

\if\stable1
  \newcommand\todo[1]{}

  \newcommand\status[1]{}
  \newcommand\defend[1]{}
\else
  \newcommand\todo[1]{\textcolor{red}{[TODO: #1]}}

  \newcommand\status[1]{\textcolor{gray}{\bf [Status: #1]}}
  \newcommand\defend[1]{\textcolor{magenta}{[Cite: #1]}}
\fi

\if\stable1

\newcommand\pl[1]{}
\newcommand\rb[1]{}
\newcommand\kc[1]{}
\newcommand\ak[1]{}

\else

\newcommand\pl[1]{\textcolor{purple}{[Percy: #1]}}
\newcommand\rb[1]{\textcolor{magenta}{[Rishi: #1]}}
\newcommand\kc[1]{\textcolor{blue}{[Katie: #1]}}
\newcommand\ak[1]{\textcolor{green}{[Ananya: #1]}}

\fi

%% file: sections/0-abstract.tex
\begin{abstract}
As the scope of machine learning broadens, we observe a recurring theme of \textit{algorithmic monoculture}: the same systems, or systems that share components (\eg training data), are deployed by multiple decision-makers. 
While sharing offers clear advantages (\eg amortizing costs), does it bear risks? 
We introduce and formalize one such risk, \textit{outcome homogenization}:
the extent to which particular individuals or groups experience negative outcomes from all decision-makers. 
If the same individuals or groups exclusively experience undesirable outcomes, this may institutionalize systemic exclusion and reinscribe social hierarchy. 
To relate algorithmic monoculture and outcome homogenization, we propose the \textit{component-sharing hypothesis}: if decision-makers share components like training data or specific models, then they will produce more homogeneous outcomes. 
We test this hypothesis on algorithmic fairness benchmarks, demonstrating that sharing training data reliably exacerbates homogenization, with individual-level effects generally exceeding group-level effects.
Further, given the dominant paradigm in AI of foundation models, \ie models that can be adapted for myriad downstream tasks, we test whether model sharing homogenizes outcomes across tasks. 
We observe mixed results: we find that for both vision and language settings, the specific methods for adapting a foundation model significantly influence the degree of outcome homogenization. 
We conclude with philosophical analyses of and societal challenges for outcome homogenization, with an eye towards implications for deployed machine learning systems.
\end{abstract}

%% file: sections/1-introduction.tex
\section{Introduction}
\label{sec:introduction}

Machine learning is built on strong traditions of sharing: we share datasets (\eg ImageNet), models (\eg BERT), libraries (\eg PyTorch), optimizers (\eg Adam), evaluations (\eg SuperGLUE) and much more.
This ethos of sharing serves the field well: we are able to repeatedly capitalize on the effort required to build high-quality assets (\eg ImageNet has supported thousands of researchers in computer vision), and improvements to these assets have sweeping benefits (\eg BERT raised all boats in NLP).
Yet does sharing also have risks? Could this core institution of the field lead to undesirable outcomes?

We observe that certain forms of sharing can be reinterpreted as monoculture: \citet{KleinbergRaghavan2021} define \textit{algorithmic monoculture} as the state "in which many decision-makers all rely on the [exact] same algorithm." 
In parts of society where algorithmic systems are ubiquitous, we see trends towards such monoculture \cite{Moore2018, Engler2021}.
Monocultures often pose serious risks: \citet{KleinbergRaghavan2021} show monoculture is suboptimal for decision-makers when their decisions are interconnected, as when they compete to hire job candidates. 
In ML, our sharing practices often are more complex than sharing the entire algorithmic system: 
should we think of our practices of sharing assets in ML as monoculture and, if so, what harms should we worry about?

We investigate this question by introducing the risk of \textit{outcome homogenization}, \ie the phenomenon of individuals (or groups) exclusively receiving negative outcomes from \textit{all} decision-makers they interact with.
For example, a job applicant may be rejected from every job they apply to due to the use of similar algorithmic resume screening systems at all companies.
Homogeneous outcomes, especially in certain high-stakes settings, constitute serious harms to individuals: someone who is rejected from every employment opportunity, or denied admission at every school, may be severely compromised (\eg unable to provide for their family, unable to secure an education).  
We view outcome homogenization as an important class of \textit{systemic} harms that arise when we study social \textit{systems}, \ie harms that require observing how individuals are treated by many decision-makers.\footnote{In fact, outcome homogenization is a systemic harm that may arise even in the absence of algorithmic monoculture, though this work is restricted to settings where monoculture is present.}
In \autoref{sec:outcome-homogenization}, we conceptually motivate outcome homogenization in the context of (algorithmic) hiring.
In \autoref{sec:formal-model}, we introduce the first mathematical formalism for outcome homogenization: we measure homogenization as the observed probability of systemic failure normalized by a base rate.

To link the social practice of sharing in ML with the social harm of outcome homogenization, we pose and test the \textit{component-sharing hypothesis}.
If decision-makers share the same underlying components, such as training data and machine learning models, then they will tend to systematically fail the same individuals or groups.
In reasoning about the sharing of components, we broaden the initial definition of algorithmic monoculture from \citet{KleinbergRaghavan2021}: monoculture is not just when decision-makers deploy the exact \textit{same} system, but also when they deploy \textit{similar} systems (here, meaning similar in how they were constructed). 
We investigate how two types of shared components --- training data and foundation models --- contribute to homogeneous outcomes.

In \autoref{sec:data-sharing}, we demonstrate that data sharing often homogenizes outcomes for individuals and for racial groups across 3 algorithmic fairness datasets.
In \autoref{sec:model-sharing}, we discuss how the rise of foundation models \citep{bommasani2021}, \ie
models that can be adapted to myriad downstream tasks, could yield unprecedented homogenization.\footnote{Gary Gensler, the current chair of the US Securities and Exchange Commission (SEC) concurrently has raised this concern: \url{https://mitsloan.mit.edu/ideas-made-to-matter/secs-gary-gensler-how-artificial-intelligence-changing-finance}.} 
Based on experiments with foundation models for vision (CLIP) and language (RoBERTa), to our surprise, we find the use of foundation models does not always exacerbate outcome homogenization.
Instead, we find the specific mechanism for adapting the foundation model to the downstream task significantly influences homogenization: for example, linear probing consistently leads to more homogeneous outcomes than finetuning for both modalities.
Through these experiments, it is clear that the relationship between sharing and homogenization is not fully explained by our hypothesis, but that there is some evidence that sharing homogenizes outcomes.
To advance the study of homogenization, we conclude with philosophical analysis through an Andersonian relational egalitarian lens (\autoref{sec:relational}), 
practical challenges for its diagnosis, measurement, and rectification (\autoref{sec:challenges}) and future directions for the research community (\autoref{sec:conclusion}). 

%% file: sections/2-outcome-homogenization.tex
\section{Outcome Homogenization in Hiring}
\label{sec:outcome-homogenization}
To illustrate outcome homogenization and its potential causes (including algorithmic monoculture), we consider the motivating example of hiring, specifically the resume screening phase. 
Companies use resumes to screen job applicants, choosing which candidates to interview and which to reject.  
Maximum homogenization occurs when every company makes the same decision about each candidate, such that each lucky candidate is interviewed by all companies and each unlucky candidate by no companies. 
We say that the unlucky candidates who receive no interviews experience a \textit{systemic failure}.\footnote{A fundamental consideration is that just outcomes in hiring are contested: the notions of merit and ground truth are much more subjective than, say, classifying images as dogs or cats \citep{kasy2021}. 
For this illustrative example, we do not delve into this, though we acknowledge that some individuals can be justifiably rejected from all opportunities (\eg those attempting to become lawyers without passing the bar exam). 
Hence, the interpretation of homogeneous outcomes will need to be contextual, and is likely to be value-laden in allocative contexts such as hiring, education, lending, and health.}

\noindent \textbf{What factors might homogenize outcomes in human decision-making?}
Even in the absence of algorithms, we observe homogeneous outcomes in many settings.
In hiring, historically, hiring managers at each company decided who to interview and often agreed in their decisions. 
This agreement can be attributed to multiple sources: first, if the needs of each company were identical, then managers at different companies may be incentivized to interview the same candidates, thereby homogenizing outcomes.
Second, if hiring managers' choices are influenced by the same social biases, they will mistakenly reject similar people (\eg those belonging to marginalized groups), thereby contributing to homogenization.
Bias in resume screening is well-documented and remains significant \citep[][\textit{inter alia}]{Jowell1970, Bertrand2004, Kline2021}. 

However, neither explanation implies that systemic failures are inevitable. 
Companies have different needs, and resumes are imperfect predictors of success in role, so the ``best" candidates will likely differ across companies.
Further, bias is not uniform across companies: \citet{Kline2021} find that 21\% of firms were responsible for 46\% of the racial bias in interview decisions. 
Even if decisions are influenced by the same group-level biases, different companies may choose different individual members of the advantaged and disadvantaged groups.
Variance in company needs, in prevalence of bias, and in individual hiring manager preferences all make it more likely that different resumes survive the screening stage at different companies, ensuring some heterogeneity/diversity in resume screening outcomes.

\noindent \textbf{How do these dynamics change with the introduction of algorithmic decision-making?}
Algorithmic resume screening is ubiquitous: many large companies deploy resume screening algorithms to parse resumes and inform/decide which applicants advance \citep{sonderling2022promise}. 
As a stylized example, if every company deploys the \textit{same} deterministic system and has the same hiring criteria, then outcomes will be necessarily homogeneous: individuals will either receive interviews at every company or be rejected by all of them (\ie systemic failure).
While this may seem unlikely (\ie different companies, especially competitors, relying on the same algorithmic system), we observe that many companies rely on third-party vendors to provide these algorithmic hiring tools.
Hence, while the status quo is likely more complex than this stylized example (\eg vendors could customize the algorithms it uses for each client, each client may integrate a vendor's recommendation with other information sources to make a decision), we do observe that a few major vendors dominate the marketplace for algorithmic resume screening (\eg 700 companies, including over 30\% of Fortune 100 companies, rely on Hirevue \cite{Hirevue2021}).
This practice of different decision-makers deploying the same system is defined as \textit{algorithmic monoculture} by \citet{KleinbergRaghavan2021}.

More generally, different companies may instead deploy \textit{similar}, but non-identical, systems. 
We expand the definition of algorithmic monoculture to encapsulate this broader setting, which is also alluded to in \citet{KleinbergRaghavan2021}.
\citet{Engler2021} describes this as the reality for college enrollment management algorithms, writing "there are a relatively small number (between 5 and 10) of prominent vendors in the enrollment management algorithm market, \dots their process and analytics are markedly similar. Since their processes seem relatively consistent, the outcomes might be as well --- potentially leading to consistently good results for students who match the historical expectations of colleges, and consistently poor results for students who don't". 

\noindent \textbf{Component-sharing hypothesis.}
In this work, we study systems that are related in how they are constructed, akin to what is described by \citet{Engler2021}.  
We pose the \textbf{component-sharing hypothesis} that relates such algorithmic monoculture with outcome homogenization:
\textit{If deployed algorithmic systems share components, outcome homogenization will increase (\ie there will be more systemic failures)}. 
In this work, we empirically test this hypothesis for two prominent forms of component sharing: (i) the sharing of training data in training all deployed systems (\autoref{sec:data-sharing}) and (ii) the sharing of the same foundation model for building all deployed systems (\autoref{sec:model-sharing}). 

%% file: sections/3-metrics.tex
\section{Formalizing Outcome Homogenization}
\label{sec:formal-model}
While prior work \citep{KleinbergRaghavan2021, creel2022} alludes to outcome homogenization, here we provide the first mathematical formalism of outcome homogenization.\footnote{The formal model of \citet{KleinbergRaghavan2021} is related, but substantially distinct. 
Concretely, their formalism considers harms experienced by decision-makers, whereas we center decision-subjects  (\ie individuals).}
In line with our running example of resume screening, we first formalize outcome homogenization for individuals in terms of \textit{systemic failures}.
We then generalize to the group setting, where groups are systemically excluded rather than individuals. 
Since homogenization is a concept we introduce in this work, having formally defined our metrics, we relate our homogenization metrics to established metrics for correlation, fairness, robustness, and accuracy. 

\subsection{Homogeneous Outcomes for Individuals}
\label{subsec:formal-individuals}
\noindent \textbf{Setup.}
Since we define outcome homogenization as a systemic phenomenon, we consider a social \textit{system} of individuals (\eg job applicants) and decision-makers (\eg employers).
In this system, we will assume that each individual $j$ interacts with every decision-maker $i$.\footnote{Generalizations are provided in \autoref{app:metrics}.}
In the context of machine learning, we will say each decision-maker $i$ is represented by a machine learning model $h^i$, though our formalism of homogenization does not depend on the nature of this model (or even that it is a model; our measures apply equally to human decision-making). 

As an example, an individual $j \in [N]$ submits features $x_j^i$ (\eg their resume) as input to company $i \in [k]$ to receive an outcome $h^i(x_j^i) = \hat{y}_j^i$ (\eg an interview). 
Let $D^i$ be the empirical distribution of inputs $x^i$ for company $i$.\footnote{Note that our framework is general: we permit the deployed models to be for different tasks and for the individual's inputs to not be the same, though in our resume screening example all the models perform the same task and applicants often submit the same resume to different companies.}

\noindent \textbf{Failures.}
Let $F^i(x_j^i)$ indicate if $\hat{y}_j^i$ is a failure, \ie individual $j$ experiences a negative outcome from model $h^i$.
The failure rate for model $h^i$ is
\begin{equation}
\label{eq:failure_rate}
    \fail(h^i) \triangleq \E_{x^i \sim D^i}F^i(x^i) = \Pr_{x^i \sim D^i}\left[F^i(x^i) = 1 \right].
\end{equation}
In classification, failures can be classification errors ($F^i(x^i) \triangleq \1\left[h^i(x^i) \neq y^i \right]$): in this case, the failure rate is simply the empirical classification error. 
In our experiments, we will operate in the classification setting, though other settings can be accommodated by this framing (\eg in hiring and education, there may not be a "ground truth"; the relevant notion of individual harm to consider may instead be the rate of rejection from opportunities~$F^i(x^i)~\triangleq~\1\left[h^i(x^i) = -1\right]$).

\noindent \textbf{Systemic failures for individuals.}
If an individual exclusively experiences failure, we say they experience \textit{systemic failure}.
The \textit{observed rate of systemic failure} $\systemic(h_1, \dots, h_k)$ is
\begin{equation}
\label{eq:systemic}
    \systemic(h_1, \dots, h_k) \triangleq \E_{j}\left[\prod_{i}F^i(x_j^i) \right]
    = \Pr_{j} \left[F^1(x_j^1)= 1 \wedge \dots \wedge F^k(x_j^k) = 1 \right].
\end{equation}

\noindent \textbf{Homogenization metric for individuals.}
$\systemic$ quantifies homogeneous outcomes, but is difficult to compare across systems with different underlying accuracies: $\systemic$ will in general be higher for less accurate systems independent of a \textit{specific} tendency to pick (\ie fail) on the same person. 
While we may sometimes want to combine accuracy and outcome homogenization into an overall measure of utility or social welfare, which $\systemic(h_1, \dots, h_k)$ implicitly does, we focus on a \textit{relative} measure of homogenization that aims to disentangle accuracy from homogenization.
Thinking about the societal impact of ML, we are interested in outcome homogenization even, and perhaps especially, when models are highly accurate.
That is, we should not neglect those individuals who are failed uniformly merely because the overall system-wide accuracy suggests a rosy picture.

As a result, we measure individual-level outcome homogenization for a social system $\left\{h^i\right\}_{i=1}^k$ by normalizing the \textit{observed} rate of systemic failure by the \textit{expected} rate of systemic failure.
\begin{equation}
\label{eq:homogenization-systemic}
    \text{\OH}(h^1, \dots, h^k) 
    \triangleq 
    \frac{\systemic(h_1, \dots, h_k)}{\prod\limits_{i} \fail(h^i)} 
    = 
    \frac{\E\limits_{j}\left[\prod\limits_{i }F^i(x_j^i)\right]}{\prod\limits_{i}\left[\E\limits_{j}F^i(x_j^i) \right]}
\end{equation}
This measure is the ratio between (i) the probability that an individual experiences systemic failure and (ii) the probability that randomly sampled outputs for each model are all failures.
That is, the measure captures how the rate of systemic failure changes when we attend to the structure of individuals. 

\subsection{Homogeneous Outcomes for Groups}
\label{subsec:formal-group}
In addition to individual-level homogenization, we also measure group-level homogenization.
While our individual-level metric individualizes harm, complementing work on group-level biases, we may also want to identify the extent to which (possibly marginalized) social groups (\eg Black women) are systemically excluded.
Further, we often lack individual-level information (\eg due to privacy concerns; see \autoref{sec:societal}), or study algorithmic deployments that do not share individuals (\eg hiring in different states).

\noindent \textbf{Groups.}
For each input $x^i$, denote the associated group as $G(x^i) \in \mathcal{G}$.
Group identity can correspond to the data producer (\eg the age of a user querying a search engine) or the data subject (\eg the race of an individual subject to face recognition).
Let $D^i_g$ be the empirical distribution of inputs for group $g$ (\ie $\{x^i \mid G(x^i) = g \}$).
The \textit{group failure rate} $\fail_g(h^i)$ is
\begin{equation}
\label{eq:group-failure-rate}
    \fail_g(h^i) \triangleq \E_{x^i \sim D^i_g}F^i(x^i).
\end{equation}

\noindent \textbf{Homogenization metric for groups.}
To measure group-level homogenization, we modify our individual-level metric: a weighted average over groups replaces the simple average over individuals.

\begin{align}
\label{eq:homogenization-group}
    \text{\OHG}(h^1, \dots, h^k) 
    &= 
    \frac{
    \mathlarger{\sum\limits_{g}} \left[W(g)\prod\limits_i \fail_g(h^i) \right]
    }
    {
    \prod\limits_i \fail(h^i)
    }   
\end{align}

\noindent \textbf{Weights.}
We consider three weighting schemes, specified by categorical probability distributions $W$ distributed over $\mathcal{G}$ (full definitions in \autoref{app:metrics-group}):
\begin{description}
\item[Average (\Havg)] $W$ weights each group proportional to its \textit{frequency} across all deployments. 

\item[Uniform (\Hunif)] $W$ is the uniform distribution, so $W(g) = \frac{1}{|\mathcal{G}|}$.
\item[Worst (\Hworst)] $W$ assigns weight $1$ to the group $g_{\textbf{worst}}$ with the highest systemic failure rate and $0$ to all other groups. 
This reduces the numerator to simply be the systemic failure rate for $g_{\textbf{worst}}$.
\end{description}

We introduce these weight functions to clarify that, much like having both individual-level metrics and group-level metrics, we may want to weight groups differently in different circumstances.
For example, weighting by frequency may provide a useful overall measurement of homogenization but obscure systemic exclusion experienced by minority groups or specifically the worst-off group. 

\subsection{Understanding Our Metrics}
\label{subsec:understanding-metrics}
As a ratio of probabilities, our metrics take values in $[0, \infty)$ where $0$ indicates no systemic failures, $1$ indicates the observed rate matches the expected rate, 
and values greater than $1$ indicate homogeneous outcomes beyond what can be expected from the underlying failure rates .
In the individual setting, we assume each individual interacts exactly once with each decision-maker, which may not hold in practice (\eg people may apply multiple times or not apply at all to a given company).
We appropriately generalize our individual-level metric to address this in \autoref{app:metrics}.
Further, in the group setting, we recover the individual-level metric using the \textbf{uniform} weighting (or the \textbf{average} weighting) if each individual's inputs are treated as belonging to their own group.

\subsection{Relationship with Other Metrics}
\label{subsec:metrics-relationship}

Since we introduce (several) metrics, we consider how they relate to metrics for related constructs (\eg accuracy, fairness, robustness).
This speaks to the convergent and divergent validity of our metrics \citep{campbell1959, messick1987, jacobs2021}, \ie whether they are adequately correlated with metrics of similar constructs and adequately uncorrelated with metrics of dissimilar constructs.
Here, we discuss theoretical relationships, whereas in \autoref{subsec:correlations} we look at the empirical correlations.

\noindent \textbf{Accuracy.} 
When failures are errors, we design our metrics to minimize (anti-)correlation with accuracy.
While not theoretically guaranteed, we empirically demonstrate this in \autoref{tab:correlations}.
With that said, we expect there will be settings where the two are correlated: our goal is not to forcibly (\ie mechanically) ensure no correlation in the technical sense, but to ensure that we do not neglect homogeneous outcomes in highly accurate systems (\ie neglect the individuals who are systemically failed even when the overall picture may seem favorable). 

\noindent \textbf{Fairness and robustness.} 
Beyond accuracy, outcome homogenization in the group setting is closely related to fairness and robustness.
However, we emphasize that outcome homogenization is fundamentally about correlated outcomes for social \textit{systems}, whereas almost all robustness or fairness metrics are defined for a single model.
Further, almost all formalisms of fairness and robustness necessitate groups (even if these groups need not be specified in advance), whereas our formalism of homogenization is attractive in that it can be meaningfully defined for singular individuals. 
Recent work \citep{zhao2019, damour2020, wang2021} has initiated the study of fairness in multi-task learning, however these works focus on favorable overall trade-offs across tasks as opposed to systemic modes of failure.
Conversely, our metrics cease to be interesting (\eg \OH~is always $1$) when there is a single decision-maker as systemic failures degenerate to single-model failures.

At a more fine-grained level, algorithmic fairness metrics \citep[\eg][]{dwork2012, hardt2016} emphasize \textit{discrepancies} between individuals/groups.
In contrast, our metrics do not (explicitly) center these differences: we are interested in the observed rate of systemic failures (and whether this exceeds the expected rate).
Performance disparities are not \textit{sufficient} for outcome homogenization: if the performance disparities across decision-makers do not align, then outcomes may not be homogeneous.
For robustness metrics, our metric \Hworst in the worst-case setting closely resembles the metrics studied in work on worst-group robustness \citep[\eg][]{sagawa2020}. 
In particular, when there is only one decision-maker, our metric recovers the standard worst-group error normalized by the overall error.

\subsection{Alternative Metrics}
\label{subsec:alternative-metrics}

In \autoref{app:metrics}, we more extensively discuss desiderata for our metric, alternatives we considered, and how we arrived at the metrics we present in the main paper.
With that said, we also note conditions where we may instead favor alternatives, as well as connections to familiar quantities like the covariance, Pearson correlation, and (pointwise) mutual information in the binary setting ($k=2$).

%% file: sections/4-data-sharing.tex
\section{Data-sharing Experiments}
\label{sec:data-sharing}

Having stated our mathematical formalism and metrics for outcome homogenization, we test if sharing training data leads to outcome homogenization.
We consider widely used algorithmic fairness datasets \citep{fabris2022}.
While demonstrating systemic failures for these specific datasets may not signify concerns of immediate/direct social consequence, these datasets do represent relevant social contexts where other forms of inequity have been documented.

\noindent \textbf{Data.}
We work with two datasets: \textbf{German Credit} \citep[\textbf{GC};][]{dua2019}, the third most widely used fairness dataset, and \textbf{ACS PUMS} \citep{ding2021}, which was built to replace the most widely used fairness dataset, \textbf{UCI Adult}.\footnote{We include results for a third dataset, \textbf{LSAC} \citep{wightman1998}, in \autoref{app:experiments}.} 
\textbf{GC} contains information on 1000 German contracts (\eg credit history, credit amount, credit risk for the individual): following \citet{wang2021}, we consider two prediction tasks of 
(i) predicting if an individual receives a good or bad loan 
and (ii) predicting whether their credit amount exceeds 2000.
\textbf{ACS PUMS} contains US Census survey data recording 286 features (\eg self-reported race and sex, occupation, average hours worked per week) for 3.6 million individuals.
\citet{ding2021} construct several prediction tasks of which we use three: (i) predict if an individual is employed, (ii) predict an individual's income normalized by the poverty threshold, and (iii) predict if an individual has health insurance. 

\noindent \textbf{Individuals and groups.}
For both datasets, we have individual-level information, hence we measure individual-level homogenization.
For \textbf{ACS PUMS}, we have self-identified racial demographic metadata across 9 US Census categories (\eg American Indian, Asian, Black/African American, White, two or more races), hence we measure group-level racial homogenization.

\noindent \textbf{Experimental design.}
To test if and how data sharing influences outcome homogenization, informally, we would like to specify settings with different amounts of shared data to test how this affects outcomes.
However, in general, it is challenging to precisely articulate what it means to "share data" and, therefore, convincingly ensure that one setting involves more sharing than the other.
To address this challenge, we present a highly controlled comparison specifying two sampling protocols for the training data: \textbf{fixed} and \textbf{disjoint}.
In the \textbf{fixed} setting, we sample $n$ points without replacement from the entire training dataset, which we use to train all of the $k$ task-specific models ($k = 2$ for \textbf{GC} and $k = 3$ for \textbf{ACS PUMS}).
In the \textbf{disjoint} setting, we sample $kn$ points without replacement that we randomly partition across the $k$ task-specific models.
In other words, in the \textbf{fixed} setting, the task-specific models \textit{share the exact same training data inputs}, whereas in the \textbf{disjoint} setting, the task-specific models \textit{share the same training distribution, but not the exact data}.
We emphasize that this is a subtle difference between the settings, but it implies  \textbf{fixed} \textbf{shares more} than \textbf{disjoint}.

Having specified the training data, we train models for each of the $k$ tasks across the model families of logistic regression, SVMs, gradient-boosted trees, and small neural networks (further details in \autoref{app:experiments}).
To account for randomness, we report results averaged over 25 trials of the experiment (\ie 5 samples of training data and 5 training runs per sample for every value of $n$ we consider). 

\input{figures/german_partition_neurips_nn.figure}
\input{figures/census_partition_neurips_logistic.figure}

\noindent \textbf{Results and analysis.}
We report results in \autoref{fig:german-partition-nn} and \autoref{fig:census-partition} to demonstrate specific phenomena, deferring the remainder of the results to \autoref{app:experiments}.
In \autoref{fig:german-partition-nn} (\textbf{left}), we see clear evidence for our hypothesis: the \textbf{fixed} setting, which by construction involves more sharing, reliably shows more homogeneity than the \textbf{disjoint} setting.
Overall, across all 3 datasets and 4 model families we consider, we find that the \textbf{fixed} setting generally leads to more homogeneous outcomes than \textbf{disjoint}, which provides evidence towards our hypothesis: the use of the same training data leads to greater outcome homogenization than the use of different (but identically distributed) training data.

However, this relationship is not perfect: we do see several instances where the degree of homogeneity is similar or even sometimes greater for the \textbf{disjoint} setting (\eg regions of the \textbf{left} subplot of \autoref{fig:census-partition}).
Therefore, data sharing alone does not fully characterize homogeneity (\eg randomness in training and instability in the number of observed systemic failures are important to consider).
Conceptually, we highlight that this is critical to the character of systemic failures: data points that are failed by all models are very sensitive and, as a result, the trends can be quite fickle even in very pure settings with very controlled sharing like the ones we consider here.

Further, the trends in the error rates for \textbf{fixed} and \textbf{disjoint} are near-identical (\textbf{right} subplot of \autoref{fig:german-partition-nn}), as we would expect given the relationship between the sampling protocols.
That is, this is clear evidence of an important qualitative consequence: by looking at accuracy alone, these settings are indistinguishable, whereas they differ considerably in terms of the number of observed systemic failures.
Our measures correctly identify discrepancies in these settings, even though the underlying error rates are the same. 
Finally, as a function of dataset size, we do not see consistent trends on the relationship between dataset scale and outcome homogenization, though we unsurprisingly see the \textbf{disjoint} and \textbf{fixed} settings converge for larger dataset sizes (since the discrepancies in the sampling become negligible due to convergence in measure).

\noindent \textbf{Picking on the same person, not just the same group.}
For \textbf{ACS PUMS} in \autoref{fig:census-partition}, we contrast the individual-level and group-level homogenization.
(Recall the \textbf{average} and \textbf{uniform} metrics are the group-level analogues of our individual-level metric.)
Outcomes are consistently more homogeneous at the individual-level than for racial groups.
In fact, group-level analysis shows little homogenization (values near $1$) and little change as a function of dataset size, whereas individual-level measurement exposes greater homogeneity and more variation (which is unsurprising since group-level quantities are more extensively aggregated).
This has significant ramifications for many works on algorithmic fairness, which only consider social groups (\eg race): these works may miss systemic failures for particular individuals that are obscured at the group-level \citep[cf.][]{Kearns2018, hashimoto2018}.
Even intersectional approaches \citep{crenshaw1989} may not suffice to surface these systemic failures, unless each intersectional "group" reduces to a single individual.
As a results, to build comprehensive and holistic accounts of algorithmic harm, we recommend an individual-centric lens as essential complements to widespread group-centric analyses.

%% file: figures/german_partition_neurips_nn.figure.tex
\begin{figure}[t]
\centering
\includegraphics[width=\linewidth]{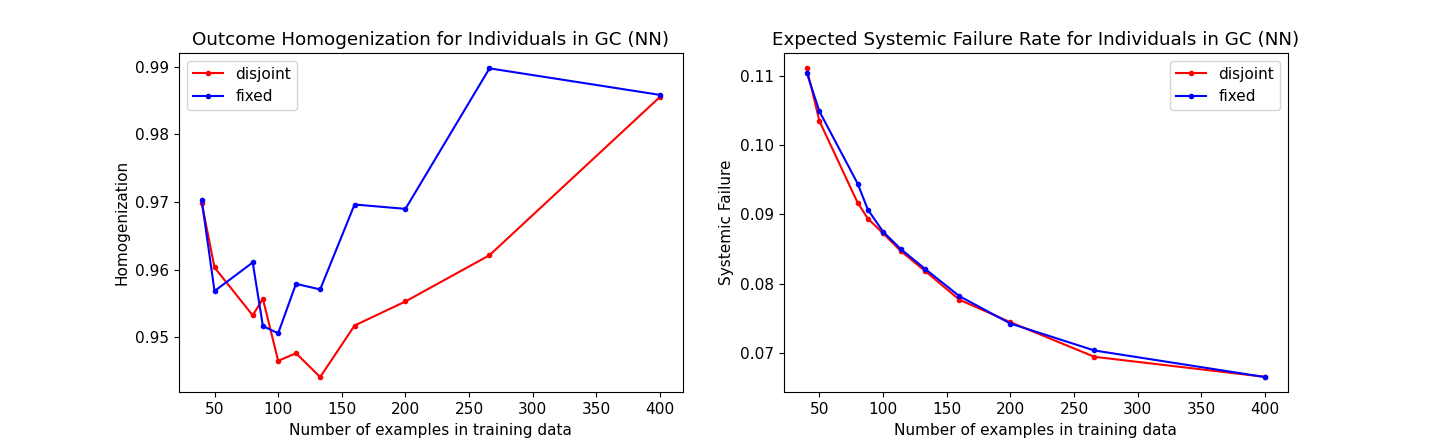}
\caption{
\textbf{Data sharing leads to more homogeneous outcomes.}
Data-sharing results for \textbf{GC} using neural network classifiers, which shows homogenization (\textbf{left}) and expected systemic failure rate (\textbf{right}) as a function of training dataset size ($x$-axis). 
}
\label{fig:german-partition-nn}
\end{figure}

%% file: figures/census_partition_neurips_logistic.figure.tex
\begin{figure}[t]
\centering
\includegraphics[width=\linewidth]{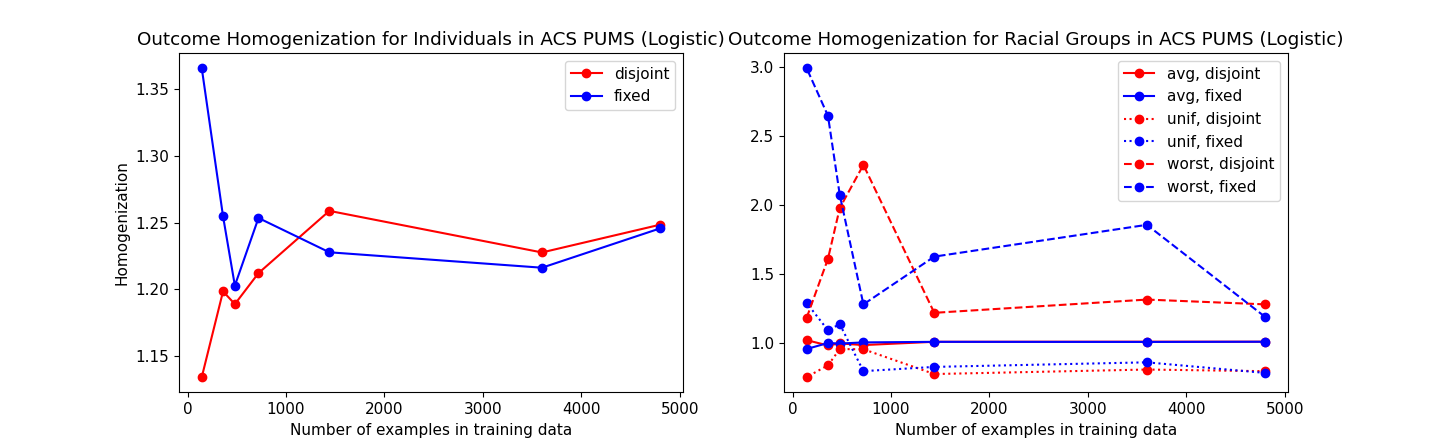}
\caption{
\textbf{Picking on the same person.}
Data-sharing results for \textbf{ACS PUMS} using logistic regression: individual-level homogenization (\textbf{left}) exceeds group-level (\textbf{right}; \textbf{avg} and \textbf{unif}).
}
\label{fig:census-partition}
\end{figure}

%% file: sections/5-model-sharing.tex
\section{Model-sharing Experiments}
\label{sec:model-sharing}

Having found that data sharing appears to exacerbate outcome homogenization, we now turn to model sharing. 
Specifically, we test how sharing \textit{foundation models} (FMs) affects outcome homogenization.
\citet{bommasani2021} define foundation models as "models trained on broad data (generally using self-supervision at scale) that can be adapted to a wide range of downstream tasks".
FMs have ushered in a sweeping paradigm shift, transforming AI research in many subareas (most notably NLP) and accelerating the productionization process from AI research to AI deployment.
Focusing on their concrete societal impact, foundation models are increasingly central to deploying ML at both startups (\eg Hugging Face, Cohere, AI21 Labs) and established technology companies (\eg Google, Microsoft, OpenAI).\footnote{The growing startup ecosystem centered on FMs has raised billions in capital as documented by \url{https://www.scalevp.com/blog/introducing-the-scale-generative-ai-index}.}

Sharing is endemic to the FM paradigm: to justify their immense resource requirements, models must be used repeatedly for costs to amortize favorably.
In the extreme, if an entire domain like NLP comes to build almost all downstream systems on one or a few FMs, then any biases or idiosyncrasies of these models that pervasively manifest downstream could potentially yield unprecedented systemic failures and outcome homogenization \citep{bommasani2021, fishman2022}.
We see initial evidence for such algorithmic monoculture: BERT was downloaded 24 million times in the past month\footnote{\url{https://huggingface.co/bert-base-uncased} as of November 2022.} alone and GPT-3 enables hundreds of deployed applications.\footnote{\url{https://openai.com/blog/gpt-3-apps/}}
Consequently, we believe it is especially timely to understand if, and to what extent, outcomes get homogenized as these models become entrenched as infrastructure.

\subsection{Experiments}
\label{subsec:fm-experiments}
\noindent \textbf{Data.}
To test how foundation models influence homogenization, we conduct experiments for both vision and language data.\footnote{Full reproducibility details for vision are in \autoref{app:vision}; for language are in \autoref{app:language}.}
On the vision side, we work with the \textbf{CelebA} dataset \citep{liu2015} of celebrity faces paired with annotations for facial attributes.
For each face image, given the associated attributes, we define two tasks (\textbf{Earrings}, \textbf{Necklace}) that involve predicting whether the individual is wearing the specific apparel item.
Attribute prediction in CelebA has been studied previously in work on fairness and robustness \citep{sagawa2020, khani2021, wang2021}.
On the language side, we use four standard English text classification datasets following \citet{gururangan2019}: \textbf{IMDB} \citep{maas2011}, \textbf{AGNews} \citep{zhang2015}, \textbf{Yahoo} \citep{chang2008}, and \textbf{HateSpeech18} \citep{gibert2018}. 

\noindent \textbf{Individuals and groups.}
Since the vision tasks are all based on CelebA, we have individual-level information.
However, since the language tasks involve entirely different data (\eg movie reviews vs. news articles), there is no (shared) individual-level information.
At the group-level, for vision we use annotations for \textit{hair color} and for whether the individual has a \textit{beard}, whereas for language we automatically group inputs by \textit{binary gender}.
We deliberately do not use gender or racial information in CelebA because we believe inference of such information from face images is fraught, and has been the subject of extensive critique. 

\noindent \textbf{Experimental design.}
To test if and how model sharing influences outcome homogenization, akin to data sharing, we must precisely specify what it means to share models. 
In the vision experiments, we produce task-specific models for each task by either (i) training from \textbf{scratch} on CelebA data, (ii) linearly \textbf{probing} by fitting a linear classifier on features from the CLIP foundation model \citep{radford2021}, or (iii) \textbf{finetuning} CLIP.
To ensure meaningful comparisons, the models trained from scratch shared the same ViT architecture \citep{dosovitskiy2021} used in CLIP but with weights initialized randomly.

In the language experiments, we further hone in on the specific \textit{adaptation method} used to adapt the foundation model (specifically RoBERTa-base \citep{liu2019}) to each task.
We consider (i) linear \textbf{probing}, (ii) \textbf{finetuning}, and (iii) \textbf{BitFit} \citep{benzaken2022}, which is a recent \textit{lightweight finetuning} method that involves freezing all the FM weights except the bias parameters which are updated as in finetuning.
Consequently, BitFit is an intermediary between probing and finetuning, which has been shown to achieve similar accuracy as finetuning while updating very few FM parameters \citep{benzaken2022}.
For both vision and for language, all models are trained for the same number of epochs and we repeat each experiment for 5 random seeds per training approach.

\noindent \textbf{Hypotheses.}
Much like data sharing, model sharing is graded and is not binary: different downstream systems can share varying degrees of underlying models.
By design, our experimental design suggests a continuum in sharing: first, downstream system either can share a foundation model or not (\textbf{scratch}).
Second, among methods that involve foundation models, all methods initialize the weights using the FM weights but differ in which parameters remain the same \textit{after} adaptation is completed: \textbf{finetuning} changes all the parameters, \textbf{BitFit} only changes the bias parameters, and \textbf{probing} changes none of the parameters.
As a result, overall, we can rank methods from most to least sharing as (i) \textbf{probing}, (ii) \textbf{BitFit}, (iii) \textbf{finetuning}, (iv) \textbf{scratch}, which leads us to predict the amount of homogenization will also follow this ranking under our component-sharing hypothesis.

\input{figures/main-results.figure}

\noindent \textbf{Results and analysis.}
In \autoref{fig:main-results} (\textbf{left}), across all vision settings, we surprisingly find that \textbf{scratch} is the most homogeneous, \ie more homogeneous than either approach involving shared FMs.
This is the opposite of what we hypothesized: we posit that this may indicate model sharing is not the key explanatory variable for outcome homogenization here, but instead it is a more complex form of data sharing.
Specifically, we conjecture that since the \textbf{scratch} models are only trained on \textbf{CelebA} data, whereas the others also are trained on the much larger WebImageText via the CLIP foundation model, this may mean that the models based on CLIP are effectively regularized from learning idiosyncrasies of \textbf{CelebA} that the \textbf{scratch} models acquire.
This may more generally suggest that a more correct hypothesis around data sharing should factor in the relationship (\eg distribution shift) between the training data and the evaluation data for each model.
Additionally, we find \textbf{probing} is consistently more homogeneous than \textbf{finetuning}, which aligns with our hypothesis.
Finally, akin to the data-sharing experiments (\autoref{sec:data-sharing}), we once again find that outcome homogenization is significantly higher for individuals than for groups (comparing to \Havg~and \Hunif). 

In \autoref{fig:main-results} (\textbf{right}), across all language settings, we find the trends in homogenization largely matches what our hypothesis predicts. 
However, instead of \textbf{BitFit} demonstrating more homogenization than \textbf{finetuning}, we find they are roughly equally homogeneous. 
In particular, based on shared parameters, the final \textbf{BitFit} models across tasks share $99.92\%$ of the parameters\footnote{Excluding the fully learned parameters for the classifier head.} from the FM weight initialization, which is much closer to \textbf{probing} ($100\%$) than to \textbf{finetuning} ($0\%$).  
That is, while these results agree with our hypothesis (\ie we see the most homogenization in the \textbf{probing} setting, much like we saw in the vision experiments), the number of shared parameters is probably not the right lens for understanding model sharing. 
Instead, we may need a more sophisticated account of how the FM weights change due to the adaptation process \citep[cf.][]{kumar2022}.
More broadly, these results do suggest parameter-sharing effects contribute to outcome homogenization within the foundation model regime, but comparisons between foundation models and no foundation models may be more complex to explain.

\subsection{Correlations between Metrics}
\label{subsec:correlations}
Since we introduce several metrics, we measure the correlations between our metrics. 
Further, we measure correlations with accuracy (specifically, the expected rate of systemic failure) to test if homogenization is disentangled from accuracy.
Since outcome homogenization is related to fairness, we also measure the correlation between our metrics and a standard group fairness metric.
Fairness metrics are generally defined for a single model $h$, whereas we study entire systems $\{h^i\}_{i=1}^k$.
We extend the unfairness definition used by \citet{khani2019} as the variance in the systemic failure rates across groups.
\begin{equation}
    \textsc{unfairness}_G(h^1, \dots, h^k) \triangleq \var\limits_{g} \left[ \prod\limits_i \fail_g(h^i) \right]
\end{equation}
\input{tables/correlations.table.tex}
\noindent \textbf{Results.}
In \autoref{tab:correlations}, we report the pairwise correlation between metric pairs, based on the models we trained in \autoref{subsec:fm-experiments}.
These correlations are for 45 systems (3 methods $\times$ 3 groupings $\times$ 5 random seeds) of 2 models for vision and 15 systems of 4 models for language. 
For vision, our metrics are highly correlated with each other, whereas for language, \Hunif~patterns quite differently (columns 1-3, 6-8).
Since the group-level metrics differ in how they weigh groups, we find this result is largely predictable: for the vision data, groups (\eg hair colors) all share similar frequencies, whereas the female group is significantly rarer than the male group in the language datasets.
For both language and vision, we find that our metrics are generally not correlated, or perhaps weakly correlated, with accuracy/error as we intended (columns 4, 9).
With respect to fairness, our worst-case metric \Hworst~is strongly correlated for both modalities, but for the other two metrics we see no linear correlations and only monotone correlations for the vision experiments (columns 5, 10).
This is in line with our broader expectations that (system-level) fairness and outcome homogenization are indeed related (especially for the worst-performing group), but that given they are distinct theoretical constructs, they should not always be correlated \citep{campbell1959}.

\subsection{Discussion}
\label{subsec:discussion}

Across our experiments, we provide considerable evidence that sharing leads to homogeneous outcomes, but that it is incomplete explanation of homogeneity.
This is particularly relevant when the findings in \autoref{sec:data-sharing} and \autoref{sec:model-sharing} are contrasted, given model sharing in the foundation model regime indirectly implies immense data sharing via the training data (as mediated by the FM weight initialization). 
We emphasize that the regimes for these findings are quite different: low-dimensional tabular data with simple model families in our data-sharing experiments vs. high-dimensional images/text with large neural networks in our model-sharing experiments, so discrepancies in the findings may be attributable to these differences.
More broadly, we believe a more complete explanation requires accounting for the data distributions and the associated distribution shifts (\eg between training and adaptation) at play.
What we believe is clear, however, is that our findings provide an empirical basis to build on our conceptual arguments that sharing in machine learning can increase homogenization.
This motivates investigation into real deployments of machine learning: for example, does sharing/monoculture lead to homogenization in algorithmic hiring?

%% file: figures/main-results.figure.tex
\begin{figure}[t]
\centering
\includegraphics[width=\linewidth]{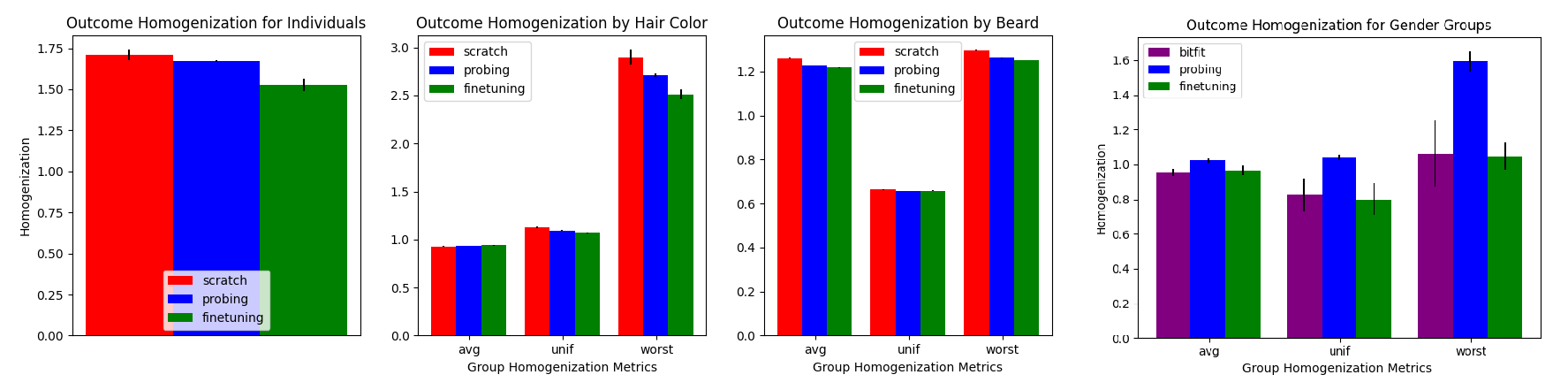}
\caption{
\textbf{Model-sharing does not reliably lead to more homogeneous outcomes.}
Model-sharing results as a function of training/adaptation method for vision (\textbf{left three}) and language (\textbf{rightmost}). \\
\textbf{Vision:} \textbf{\textcolor{red}{scratch}} is the most homogeneous, then \textbf{\textcolor{blue}{probing}}, then \textbf{\textcolor{ForestGreen}{finetuning}}.  \\
\textbf{Language:} \textbf{\textcolor{blue}{probing}} is the most homogeneous; \textbf{\textcolor{ForestGreen}{finetuning}} and \textbf{\textcolor{violet}{BitFit}} are similarly homogeneous.
}
\label{fig:main-results}
\end{figure}

%% file: tables/correlations.table.tex
\setlength{\tabcolsep}{3.3pt}
\begin{table*}[htp]
\resizebox{\textwidth}{!}{
\begin{tabular}{r|ccccc|ccccc}
\toprule
 & \multicolumn{5}{c|}{\textbf{Vision}} & \multicolumn{5}{c}{\textbf{Language}} \\
 & \Havg & \Hunif & \Hworst & Accuracy & Unfairness & \Havg & \Hunif & \Hworst & Accuracy & Unfairness \ \\
\midrule


\Havg & - & (\textit{0.87}, \textit{0.93}) & (0.0, \textit{0.96}) & (0.0, 0.09*) & (0.0, \textit{0.8}) & - & (\textit{0.22}, \textit{-0.47}) & (\textit{0.11}, \textit{0.56}) & (0.06*, -0.22*) & (0.02, 0.09) \\

\Hunif & (\textit{0.87}, \textit{0.93}) & - & (0.0, \textit{0.96}) & (0.0, -0.02) & (0.0, \textit{0.74}) & (\textit{0.22}, \textit{-0.47}) & - & (\textit{0.63}, \textit{-0.53}) & (0.0, 0.19*) & (0.0, -0.01) \\

\Hworst & (0.0, \textit{0.96}) & (0.0, \textit{0.96}) & - & (\textit{0.05}, 0.1*) & (\textit{1.0}, \textit{0.82}) & (\textit{0.11}, \textit{0.56}) & (\textit{0.63}, \textit{-0.53}) & - & (0.02, 0.13) & (\textit{0.13}, \textit{0.47}) \\
\bottomrule
\end{tabular}}
\caption{\textbf{Inter-metric correlations.} Correlations amidst our metrics as well as with other metrics reported as (Pearson $R^2$, Spearman $\rho$) with * significant at $p = 0.05$ and \textit{italics} significant at $p = 0.001$.}
\label{tab:correlations}
\end{table*}

%% file: sections/6-societal-considerations.tex
\section{Societal Considerations}
\label{sec:societal}

To situate our work in a broader social context, we articulate how we reason about the harms of outcome homogenization as well as identify core challenges to \textbf{diagnosing}, \textbf{measuring}, and \textbf{rectifying} outcome homogenization in real deployed systems.

\subsection{Why are Homogeneous Outcomes Harmful?}
\label{sec:relational}
To this point, we have argued homogeneous outcomes are a class of systemic harms and, in particular, a class of harms we should be paying greater attention to given pervasive practices of sharing and monoculture.
In some settings, it may feel clear that homogeneous outcomes are harmful: we surely do not want individuals to be failed by every classifier they interact with and it is at least possible for the classifiers to correctly classify the individual. 
In others, such as our motivating example of hiring, it may feel less clear.
Surely there is harm if someone is locked out from employment across all employers, especially if this lockout can be traced to a shared algorithm being deployed by each employer?
On the other hand, some people must be rejected from every employer (in the absence of broader structural change) if the total number of candidates exceeds the number of open positions, and there may be justifiable circumstances to reject someone from all positions even when there are unfilled vacancies.

Overall, understanding when homogeneous outcomes constitute a social harm is \textbf{contextual}: the particular circumstances determine how to interpret whether, and to what extent, homogenization is of moral concern.
To provide one account of how to reason about these harms, we follow Elizabeth Anderson  in taking a relational egalitarian approach \citep{Anderson1999, Anderson2016} to evaluate the social harm of homogenization. 
Relational egalitarianism, as a theory of justice, argues that individuals must \textit{relate} to each other as equals in a just society.
Relational approaches have seen recent adoption in relation to AI: \citet{birhane2021algorithmic} presents a relational account of algorithmic injustice with similar approaches taken to data governance, decision-making, and fairness \citep{viljoen2021, kasy2021, fish2022}. 

To ground a relational analysis, we return to the context of hiring.
While individual organizations may establish rankings of candidates, and indeed we would expect that companies within a market sector will often agree on a hierarchy, the same hierarchy should not consistently dominate an entire sector or territory such that some people are entirely excluded from work \citep[74]{Anderson1999}. 
Anderson argues that ``to be capable of functioning as an equal participant in a system of cooperative production requires \dots access to the education needed to develop one’s talents, freedom of occupational choice, the right to make contracts and enter into cooperative agreements with others, the right to receive fair value for one’s labor, and recognition by others of one’s productive contributions'' \citep[318]{Anderson2016}. 
If some people are consistently excluded from job interviews and, therefore, employment, they will not enjoy freedom of occupational choice; if they are excluded from higher education, they will struggle to develop their talents. 
Not only do those excluded personally suffer from the establishment of the hierarchy, they also are unable to function as equal participants in society. 
Because employment, education, and credit are foundational social goods, consistent exclusion from them risks establishing a social hierarchy of esteem or domination. 
The hierarchy of esteem in turn damages the ability of the excluded to relate to others as equal democratic citizens \citep{Anderson1999}.

Under this account, we emphasize how the moral importance of the harm depends on the scale of exclusion.
That is, in its purest form, there is a strong \textit{threshold} effect: homogeneous outcomes are most severe when the individual is denied from \textit{all} sources of employment (or education or credit or so on). 
If autonomy is access to a sufficient range of sufficiently varied opportunities, then it is not a harm to be denied one opportunity, such as a job or a loan \citep{Raz1988}. 
It is a harm, however, to be shut out of all opportunities.
For this reason, in this work, we study the extreme case in which individuals are shut out of all opportunities.\footnote{We do note that while \textit{all} might be a sufficient condition for harm, it may suffice in some contexts to be denied access to a significant fraction of opportunities. For this reason, while we study the strongest form of exclusion, we encourage future work to put forth technical, experimental, and moral analyses of homogeneous outcomes in the more general setting where individuals are denied \textit{most} opportunities, not just all opportunities.} 

\subsection{Practical Challenges for Outcome Homogenization}
\label{sec:challenges}
\noindent \textbf{Diagnosis.} 
In our work, we posit monoculture yields homogenization: to follow this approach would require knowing which deployments rely on the same vendor, dataset, or foundation model (\ie knowing where there is monoculture).
Unfortunately, how algorithmic systems are constructed is often so opaque that identifying shared components is nigh impossible.
However, if high homogenization were demonstrated, the measurement itself could justify provisions for increased transparency to identify the latent monoculture (\ie the anti-causal direction).
This provides a plausible mechanism for empowering auditors to be granted conditional access to otherwise inaccessible proprietary systems. 

\noindent \textbf{Measurement.}
Measuring homogenization only requires black box access, which is often achievable in practice \citep[see][]{buolamwini2018, raji2019, metaxa2021}.
However, identifying individual-level effects requires linking individual outcomes across deployments.
Due to privacy constraints, linking individuals across different deployments may be challenging or impossible, which motivates group-level homogenization as more generally accessible (see \autoref{subsec:formal-group}). 

\noindent \textbf{Rectification.} 
Even once outcome homogenization is identified, organizations may not be incentivized to reduce it.
In fact, homogenization neither is attributable to any single entity nor can it always be addressed by unilateral action from a single organization. 
In the face of misaligned incentives and collective action problems, regulation, policy, or other compliance mechanisms may be required. 
Potential trade-offs between organization incentives and homogenization are further complicated if the harms of homogeneous outcomes take time to observe/accrue, but the benefits of, say, maximizing accuracy are immediate.
More optimistically, \citet{KleinbergRaghavan2021} show (under specific conditions) no trade-off exists between accuracy-maximizing policies and diversifying outcomes for societal benefit.

%% file: sections/7-conclusion.tex
\section{Limitations and Conclusion}
\label{sec:conclusion}
\looseness=-1
We have introduced, formalized, and measured outcome homogenization as a systemic harm that may arise from practices of sharing in ML.
Outcome homogenization is a new, understudied, and conceptually compelling topic: its definition, interpretation, statistical estimation, mitigation, and connections to monoculture remain poorly understood in spite of this work.
We encourage future work to push in all of these directions.
As to our measure, direct optimization may not lead to desirable outcomes, potentially even contributing to ethics-washing: its interpretation must be \textbf{contextual} since the implications of homogeneous outcomes heavily depend on broader societal context.

\looseness=-1
We believe homogenization is essential to holistically characterizing algorithmic harm, especially given growing monoculture (\eg via foundation models).
Without scrutiny, its harms may insidiously entrench.
Consequently, we believe early intervention is necessary to prevent such harms in society.

%% file: sections/12-acknowledgements.tex
\section*{Acknowledgements}
The authors would like to thank Simran Arora, Sarah Bana, Zachary Bleemer, Liam Kofi Bright, Erik Brynjolfsson, Steven Cao, Niladri Chatterji, Lingjiao Chen, Roger Creel, Dora Demszky, Moussa Doumbouya, Yann Dubois, Iason Gabriel, Tatsu Hashimoto, John Hewitt, Dan Ho, Sidd Karamcheti, Pang Wei Koh, Rohith Kuditipudi, Mina Lee, Isabelle Levent, Lisa Li, Nelson Liu, Sandra Luksic, Chris Manning, Charlie Marx, Kathleen Nichols, Joon Park, Deb Raji, Rob Reich, Omer Reingold, Roshni Sahoo, Judy Shen, Mirac Suzgun, Rohan Taori, Connor Toups, John Thickstun, Shibani Santurkar, Rose Wang, Michael Xie, Michi Yasunaga, Kaitlyn Zhou, and James Zou for helpful discussions.
In addition, the authors would like to thank the Stanford Center for Research on Foundation Models (CRFM) and Institute for Human-Centered Artificial Intelligence (HAI) for providing the ideal home for conducting this interdisciplinary research.
RB was supported by the NSF Graduate Research Fellowship Program under grant number DGE-1655618.
This work was partially supported by an Open Philanthropy Project Award and a Stanford HAI/Microsoft Azure cloud credit grant. 

%% file: sections/checklist.tex
\section*{Checklist}
\begin{enumerate}
\item For all authors...
\begin{enumerate}
  \item Do the main claims made in the abstract and introduction accurately reflect the paper's contributions and scope?
    \answerYes{} See metrics section (\autoref{sec:formal-model}) and experimental sections (\autoref{sec:data-sharing}, \autoref{sec:model-sharing}).
  \item Did you describe the limitations of your work?
    \answerYes{} See metrics section (\autoref{sec:formal-model}), societal considerations (\autoref{sec:societal}), and conclusion (\autoref{sec:conclusion}). 
  \item Did you discuss any potential negative societal impacts of your work?
    \answerYes{} See societal considerations (\autoref{sec:societal}) and conclusion (\autoref{sec:conclusion}). 
  \item Have you read the ethics review guidelines and ensured that your paper conforms to them?
    \answerYes{}
\end{enumerate}

\item If you are including theoretical results...
\begin{enumerate}
  \item Did you state the full set of assumptions of all theoretical results?
    \answerNA{}
        \item Did you include complete proofs of all theoretical results?
    \answerNA{}
\end{enumerate}

\item If you ran experiments...
\begin{enumerate}
  \item Did you include the code, data, and instructions needed to reproduce the main experimental results (either in the supplemental material or as a URL)?
    \answerYes{} See \url{https://worksheets.codalab.org/worksheets/0x807c29f8eb574d1fba8f429ec78b5d1b} and \autoref{app:reproducibility}. 
  \item Did you specify all the training details (e.g., data splits, hyperparameters, how they were chosen)?
    \answerYes{} See \autoref{app:reproducibility}.
        \item Did you report error bars (e.g., with respect to the random seed after running experiments multiple times)? 
    \answerYes{} See \autoref{fig:main-results}. 
        \item Did you include the total amount of compute and the type of resources used (e.g., type of GPUs, internal cluster, or cloud provider)? 
    \answerYes{} See \autoref{app:reproducibility}. 
\end{enumerate}

\item If you are using existing assets (e.g., code, data, models) or curating/releasing new assets...
\begin{enumerate}
  \item If your work uses existing assets, did you cite the creators?
    \answerYes{} See \autoref{sec:data-sharing}, \autoref{sec:model-sharing}, and \autoref{app:reproducibility}. 
  \item Did you mention the license of the assets?
    \answerNo{} We provide references directly to where we sourced data in \autoref{app:reproducibility}.
  \item Did you include any new assets either in the supplemental material or as a URL?
    \answerNA{}
  \item Did you discuss whether and how consent was obtained from people whose data you're using/curating?
    \answerNo{} We defer to the discussions in prior work, though we did check ourselves for any specific concerns of consent and found none in our initial cursory investigation.
  \item Did you discuss whether the data you are using/curating contains personally identifiable information or offensive content?
    \answerYes{} Yes, some of our data contains PII (e.g. Census information, face images) and one of our datasets is related to hate speech detection, so this is by design. These are well-established datasets, with the respective works providing discussion on these fronts, and we do not foresee any specific harms from our usage. 
\end{enumerate}

\item If you used crowdsourcing or conducted research with human subjects...
\begin{enumerate}
  \item Did you include the full text of instructions given to participants and screenshots, if applicable?
    \answerNA{}
  \item Did you describe any potential participant risks, with links to Institutional Review Board (IRB) approvals, if applicable?
    \answerNA{}
  \item Did you include the estimated hourly wage paid to participants and the total amount spent on participant compensation?
    \answerNA{}
\end{enumerate}

\end{enumerate}

%% file: appendices/metrics.tex
\section{Homogenization Metrics}
\label{app:metrics}

\subsection{Group Homogenization Metrics}
\label{app:metrics-group}
In \autoref{subsec:formal-group}, we introduced our group level homogenization metrics.
Specifically, we note the design decision of how to weight groups and the three weightings we consider: \textbf{average}, \textbf{uniform}, and \textbf{worst}.
Here, we provide the full mathematical definition of these metrics.

For convenience, let the frequency of group $g$ in a specific dataset $D^i$ by denoted as $p^i(g)$, and the joint probability of the group across all datasets be denoted as $p(g) \triangleq \prod_{i=1}^k p^i(g)$.

\begin{align}
    \text{\Havg}(h^1, \dots, h^k) 
    &\triangleq 
    \frac{
    \sum\limits_g \left[\frac{p(g)}{\sum_{g'} p(g')}\prod\limits_i \fail_g(h^i) \right]
    }
    {
     \prod\limits_i \fail(h^i)
    }  \\
    \text{\Hunif}(h^1, \dots, h^k) 
    &\triangleq 
    \frac{
    \E\limits_g \left[\prod\limits_i \fail_g(h^i) \right]
    }
    {
     \prod\limits_i \fail(h^i)
    }  \\
    \text{\Hworst}(h^1, \dots, h^k) 
    &\triangleq 
    \frac{
    \max\limits_g \left[\prod\limits_i \fail_g(h^i) \right]
    }
    {
     \prod\limits_i \fail(h^i)
    } 
\end{align}

\subsection{Relating Individual and Group Homogenization}
\label{app:metrics-example}

In \autoref{sec:data-sharing}, we demonstrate empirically that outcomes can be more homogeneous for individuals than for racial groups.
One may be led to believe that homogenization is necessarily greater when considering finer-grained groupings since individuals (when viewed as singleton groups) are subsets of groups (\ie when grouping by individuals, no data points from different races appear in the same group as each group is a singleton). 
However, this is not true.

To elucidate this, we provide two toy scenarios that demonstrate circumstances where individual-level outcome homogenization is greater than, and is less than, group-level outcome homogenization.
(Of course, they can also be equal.)
In both settings, we will have two models (\ie two decision-makers) and two groups, where each group is comprised of two individuals.
As a result, \Havg = \Hunif, so we can compare either to \OH. 

In both settings, we will say Alice and Angelique are members of Group 1 and Bob and Bernardo are members of Group 2.

\noindent \textbf{Scenario 1.}
Let Alice and Bob be misclassified by $h^1$ but correctly classified by $h^2$.  
Let Angelique and Bernardo be misclassified by $h^2$ but correctly classified by $h^1$.
No one is misclassified by both models, hence the number of observed systemic failures is 0 at the individual level, hence \OH = 0.

However, since there is a failure within Group 1 for both models (and for Group 2 as well), the number of systemic failures is nonzero at the group level, hence $\text{\Havg}~=~\text{\Hunif}~> 0$.
Thus, in this scenario, we have seen that individual-level outcome homogenization can be less than group-level outcome homogenization.

\noindent \textbf{Scenario 2.}
Let Alice and Bob be misclassified by both $h^1$ and $h^2$.
Let Angelique and Bernardo be correctly classified by both $h^1$ and $h^2$.
At the individual level, there are 2 observed systemic failures, so the observed rate of systemic failure is $\frac{2}{4} = 0.5$.
The overall error rate for each application is 0.5, so the expected rate of systemic failure is $0.5 \times 0.5 = 0.25$.
Therefore, \OH = $\frac{0.5}{0.25} = 2$.

At the group level, the rate of systemic failures is $0.25$ for both groups.
The overall error rate for each application is still 0.5, so the denominator is also $0.5 \times 0.5 = 0.25$.
Therefore, $\text{\Havg}= \text{\Hunif} = \text{\Hworst} = 1$.
Thus, in this scenario, we have seen that individual-level outcome homogenization can be greater than group-level outcome homogenization. 

\subsection{Generalizing Individual-Level Metric}
\label{app:generalizing-individual}

In \autoref{subsec:understanding-metrics}, we note that our individual-level framing assumes that every individual $j$ produces inputs $x_j^i$ for every decision-maker $i$.
The formalism in the main paper already permits these inputs to be different across decision-makers for the same individual (\eg Bob may submit different resumes when applying to Microsoft and Google).
However, the formalism does not support two further general concepts: (i) multiple inputs per decision-maker and (ii) no inputs for some decision-maker (\eg Bob does not apply to Amazon).
To accommodate the former, we note that the notion of failure can be modified depending on how the outcomes for the multiple inputs should be aggregated (\eg a failure of a search engine may be determined by some fraction of search queries producing poor results for the user).

To address the latter concern, we introduce notation $c_j$ to indicate the subset of decision-makers that individual $j$ interacts with, \ie $c_j \subseteq \{1, \dots, k\}$.
That is, any decision-makers $i \in \{1, \dots, k\}$ that are not in $c_j$ are those that individual $j$ does not interact with.
Accordingly, we definite \OH~as: 
\begin{equation}
\label{eq:generalized-individual}
\text{\OH}(h^1, \dots, h^k) \triangleq 
\frac
    {
    \E\limits_{j}\left[\prod\limits_{i \in c_j}F^i(x_j^i) \right]
    }
    {
    \E\limits_{j}\left[\prod\limits_{i \in c_j}\fail(h^i)\right]
    }
\end{equation}
Notably, when $\forall j,~c_j = [k]$, the denominator simplifies to $\prod\limits_{i \in [k]}\fail(h^i)$, which matches \autoref{eq:homogenization-systemic}.

\subsection{Alternative Metrics}
\label{app:alternative-metrics}

In \autoref{sec:formal-model}, we introduce the metrics we use to quantify outcome homogenization.
Of course, much like the many mathematical expressions that have been used to measure bias and fairness, there are many ways to reasonably measure homogenization.
Fundamentally, given the underlying construct of outcome homogenization is largely new, we begin by recognizing our understanding of the concept is incomplete and likely will require study in real systems to truly identify the precise desiderata for a measure.

In the interim, it is difficult to assess if the metric has \textit{structural fidelity} \citep{loevinger1957}, \ie does the metric's structure faithfully captures outcome homogenization?
Further, it is unclear if the metric has sufficient predictive validity to predict long-term outcomes (\eg longitudinal harms arising from outcome homogenization) or how useful it is for testing specific scientific and social hypotheses \citep{jacobs2021}.
Ultimately, we believe the key test for the metric will be its \textit{consequential validity} \citep{messick1987}: will the metric yield positive social impact as it "both reflects structure in the world and imposes structure upon the world" \citep{hand2016}.

To facilitate understanding, we transparently discuss other metrics we considered and why they may be preferable in some circumstances. 
Ultimately, we worked with the metrics we describe in the paper as we found them to be the simplest and preferred their probabilistic interpretations, but we include reasons to prefer alternatives as we describe them.

\subsubsection{Alternative Metrics in the Binary Setting}
\noindent \textbf{Covariance and Pointwise Mutual Information.}
When $k=2$, \ie there are two decision-makers, we note that our metric bears a very close resemblance to the \textit{covariance} between the (indicator) random variables $F^1$ and $F^2$. 
In particular, the covariance is the difference of the quantities that  define the ratio for our homogenization metric.
Similarly, our metric is the \textit{pointwise mutual information} (PMI) evaluated at $(1, 1)$ up to the $\log$.
\begin{align}
\label{eq:covariance}
\text{\OH}(h^1, h^2) &= \frac{\E\limits_{j} \big[F^1 F^2 \big]}{\E\limits_j\left[F^1\right]\E\limits_j\left[F^2\right]} \\
\cov(F^1, F^2) &= \E\limits_{j} \big[F^1 F^2 \big] -  \E\limits_j\left[F^1\right]\E\limits_j\left[F^2\right]\\
\text{PMI}(F^1 = 1, F^2 = 1) &= \log\left(\frac{\E\limits_{j} \big[F^1 F^2 \big]}{\E\limits_j\left[F^1\right]\E\limits_j\left[F^2\right]} \right) = \log \left(\text{\OH}(h^1, h^2)\right)
\end{align}
With respect to the covariance, we prefer that our metric is more naturally comparable across settings where the failure rates of social systems vary, whereas the covariance is more directly tied to the absolute scale of the failure rates.
With respect to the pointwise mutual information, we note that we are simply looking at the behavior of the social system in a special case where all models fail (which is one of the $2^k$ possible outcomes an individual could receive overall), whereas the overall PMI considers all of them and is invariant to symmetries that are significant in our setting.
Further, both are traditionally studied in the binary setting, whereas we study behavior in settings where $k > 2$.

\noindent \textbf{Pearson Correlation.}
Building on the relationship with the covariance, we note that our metric therefore also resembles the \textit{Pearson correlation}.
\begin{align}
\text{\OH}(h^1, h^2) &= \frac{\E \big[F^1 F^2 \big]}{\E\left[F^1\right]\E\left[F^2\right]} \\
\text{Corr}(F^1, F^2) &= \frac{\E \big[F^1 F^2 \big] -  \E\left[F^1\right]\E\left[F^2\right]}{\sqrt{\left(\E\left[F^1\right](1 - \E\left[F^1\right])\right)\left(\E\left[F^2\right](1 - \E\left[F^2\right])\right)}}
\end{align}
In particular, when dealing with accurate models that are homogeneous (\ie~$\E \big[F^1 F^2 \big] >>  \E\left[F^1\right]\E\left[F^2\right]$), the Pearson correlation coefficient approximates our metric up to the square root in the denominator.
Arguments can be made in favor and against this square root (and more generally a $k$-th root for $k > 2$); for simplicity we favor our metric that does not introduce such normalization but acknowledge this normalization may prove to be more favorable as the metric is further stress-tested.

\subsubsection{Alternative Metrics beyond the Binary Setting}
As we note in Footnote 2, our formalism and metrics are designed to be general, meaning that they can accommodate settings where the models $h^i$ correspond to different tasks or scenarios. 
(We make use of this generality in our experiments (\autoref{sec:model-sharing}) for vision and, especially, language.)
To permit this generality, we (reductively) binarize outcomes as either failures or not in our use of the indicator functions $F^i$.
In particular, for arbitrarily different tasks, the outcome spaces and their consequences on individuals may not be (easily) related.

In some settings, specifically those where the tasks that constitute the social system are sufficiently similar, we may instead prefer a more graded \textit{loss} in the place of the binary notion of failures.
For each deployment by decision-maker $i$, denote the associated loss function as $\mathcal{L}^i$ such that $\mathcal{L}^i(h^i(x_j^i), y_j^i) = \ell_j^i$ is the loss experienced by individual $j$ when interacting with model $h^i$.
In these settings, where the loss achieved across different applications is comparable, we can consider additional measures for homogenization in terms of this loss.
\begin{align}
    \text{\OH}(h^1, \dots, h^k) 
    &= 
    \frac{\E\limits_{j}\left[\prod\limits_{i }~\ell_j^i\right]}{\prod\limits_{i}\left[\E\limits_{j}~\ell_j^i \right]} \\
    \text{MinExp}(h^1, \dots, h^k) 
    &= 
    \frac{\E\limits_{j}\left[\min\limits_{i }~\ell_j^i\right]}{\min\limits_{i}\left[\E\limits_{j}~\ell_j^i \right]} \\
    \text{ExpExp}(h^1, \dots, h^k) 
    &= 
    \frac{\E\limits_{j}\left[\min\limits_{i }~\ell_j^i\right]}{\E\limits_{i}\left[\E\limits_{j}~\ell_j^i \right]} \\
\end{align}

In words, the MinExp definition is the ratio of the average best-case loss for individuals with the loss of the best model $h^{\text{best}}$ and the ExpExp definition is the same but the denominator is the average loss of the models rather than the best loss.
These definitions bear close resemblance to the MaxMin and lexicographic (leximax and leximin) fairness definitions studied in the fairness literature \citep{dubins1961, henzinger2022}; the group-weighted analogue that use the \textbf{worst} group weighting recovers the MaxMin definition in the denominator.
(Note that the naming conventions are reversed since we define metrics in terms of loss whereas work in fairness and equitable allocations generally defines metrics in terms of utility.)

When the loss is the 0-1 classification loss, the MinExp definition and \OH~ are very similar: the numerators are the same (as the product of indicator variables is the same as their minimum) and the denominators are precisely $z = \frac{\prod_{i=1}^k \fail(h^i)}{\fail(h^{\text{best}})}$ factors of each other (\ie the failure rate of all of the models except the best one). 
Consequently, the MinExp yields values in the range $[0, 1]$ whereas \OH~is in $[0, \infty)$. 
Independent behavior across models in \OH~is guaranteed to be 1 and maximal systemic failure in MinExp is guaranteed to be 1; symmetrically, independent behavior in MinExp is $z$ (which is non-constant) and maximal systemic failure in \OH~is $\frac{1}{z}$.

More broadly, the indicator variables we use in the main paper can be seen as a special case when using the 0-1 classification loss whereas arbitrary loss functions can be mapped to indicators by thresholding the loss (which does not require the loss to comparable, as different thresholds can be applied for different deployments). 
As a result, both \OH~and MinExp have clear merits; we believe MinExp may especially be a more natural definition where individuals have \textit{choices} on which model $h^i$ to interact with of the $k$ possible models.
In these settings (\eg picking which voice assistant to use, or more generally consumer products), the losses will naturally be comparable and it may suffice for the individual to have one good option, which is more smoothly encoded by the minimum of the loss rather than the product of the indicators.

We encourage future work to explore whether the MinExp or \OH~is preferable: in many settings we expect they will be strongly correlated given they are scalings of each other that depend not on the correlated nature of errors but the overall error rates, but they may be some settings where they diverge and one is clearly preferable to the other.
Currently, we recommend work to also consider MinExp when the losses are comparable, but to default to \OH~since this is not required and \OH~has a simple probabilistic interpretation. 

%% file: appendices/reproducibility.tex
\section{Reproducibility}
\label{app:reproducibility}
All code, data, and experiments are available on GitHub and CodaLab Worksheets.\footnote{\url{https://worksheets.codalab.org/worksheets/0x807c29f8eb574d1fba8f429ec78b5d1b}}
We provide further experimental details below. 

\subsection{Census Experiments}
\label{app:census}

\subsubsection{Data}
\label{app:census-data}
We work with the \textbf{ACS PUMS} data introduced by \citet{ding2021}, which contains US Census survey data for 3.6 million individuals.
\citet{ding2021} introduce the dataset to facilitate research into algorithmic fairness and the measurement of harms associated with algorithmic systems. 
For each individual in the Census, 286 features are recorded (\eg self-reported race and sex, occupation, average hours worked per week, marital status, healthcare status).
\citet{ding2021} construct several classification tasks using this data, where each task uses some of an individual's features as inputs and one of their features as the label for the prediction task (\eg predicting if an individual's income exceeds \$50000).
Of these tasks, we work with three in particular:  \textbf{ACSEmployment} (predict if an individual is employed), \textbf{ACSIncomePovertyRatio} (predict an individual's income normalized by the poverty threshold), and \textbf{ACSHealthInsurance} (predict if an individual has health insurance). 
Following \citet{ding2021}, we split the data 80/20 for train/test. 
We access this data through their \texttt{folktables} package.\footnote{\url{https://github.com/zykls/folktables}}
We use the data for all of the US (they provided state-level data as well) from 2018, which is the primary data they analyze in their paper.
We select these tasks because \citet{ding2021} do not impose any filtering constraints in selecting data for these tasks, meaning all three tasks are posed for the same underlying individuals.
See \citet{ding2021} for more details.
License information is provided at \url{https://github.com/zykls/folktables#license-and-terms-of-use} and our use of the dataset adheres to these terms of service.
The data clearly can be personally identifying given it has census records for particular individuals, but there is no offensive content. 

\subsubsection{Models}
\label{app:census-models}

Following \citet{ding2021}, we train logistic regression models using \texttt{sk-learn} \citep{pedregosa2011} with default hyperparameters.
As a sanity check, we compared average model performance and saw it matched/exceeded what is reported in \citet{ding2021}.
For each setting (\textbf{fixed}, \textbf{disjoint}) and amount of training data, we trained 25 models across 5 random subsamples of training and 5 random seeds for model run per subsample, for each of the three tasks.
In aggregate, all of the models we trained took approximately 10 hours across 5 NVIDIA Titan Xp GPUs (or 50 hours on 1 NVIDIA Titan Xp GPU), with additional experiments/debugging that is unreported in the paper taking approximately an additional 400 NVIDIA Titan Xp GPU hours.

\subsubsection{Groupings}
\label{app:census-groupings}

We consider racial groups, which are already provided in the dataset based on the self-identified category individuals chose in providing their information to the US Census.
The specific racial categories used are: White alone, Black/African American alone, American Indian alone, Alaska Native alone, American Indian and Alaska Native, Asian alone, Native Hawaiian and Other Pacific Islander alone, other unspecified race, two or more races.

\subsection{Vision Experiments}
\label{app:vision}

\subsubsection{Data}
\label{app:vision-data}
We work with the \textbf{CelebA} dataset \citep{liu2015}, which is a widely used dataset of celebrity faces paired with annotations for facial attributes.
For each face image, given the associated attributes, we define two tasks (\textbf{Earrings}, \textbf{Necklace}) that involve predicting whether the individual is wearing the specific apparel item.
Attribute prediction in CelebA has been studied previously in work on fairness and robustness \citep{sagawa2020, khani2021, wang2021}.
Given recent documentation of significant issues with computer vision datasets \citep[\eg][]{birhane2021, birhane2021b}, we emphasize that we use the dataset solely for analytic reasons to study homogenization.
Further, given works like GenderShades \citep{buolamwini2018} that highlight the harms of face recognition, 
we emphasize that we do not use the dataset for face recognition, but instead consider apparel prediction tasks where each apparel item/accessory is clearly observable in the face image.
In addition to \textbf{Earrings} and \textbf{Necklace}, the dataset also contains attributes for \textbf{Eyeglasses} and \textbf{Neckties}.
We initially included these tasks, but observed no individual was misclassified for all four tasks since these two tasks were very easy and models rarely produced any errors (\eg the error rate for \textbf{Eyeglasses} was generally less than $1\%$).
To be able to present non-trivial results for individual-level outcome homogenization, we therefore removed these tasks from consideration so that a nonzero number of systemic failures could be observed. 
We downloaded the CelebA data from \url{https://mmlab.ie.cuhk.edu.hk/projects/CelebA.html}.
We resized images to 224-by-224, and then apply the same augmentations as in CLIP \citep{radford2021}, before feeding the image into the ViT-B/16 (which is what \citet{radford2021} do as well).
License information is provided here: \url{https://mmlab.ie.cuhk.edu.hk/projects/CelebA.html}. 
Our use of the dataset is consistent with the requirements for non-commercial research use. 
The data clearly can be personally identifying given it has face images for particular individuals, but there is no offensive content. 

\subsubsection{Models}
\label{app:vision-models}

Following \citet{radford2021}, we either use their released 150M parameter ViT-B/16 CLIP model (the largest publicly available CLIP model at the time of writing) or a randomly initialized model with the same architecture.
We used code from the official CLIP repository at \url{https://github.com/openai/CLIP}.
Since \citet{radford2021} modify the standard Vision Transformer \citep{dosovitskiy2021}, we use their modified version for our \textit{scratch} models.
As a sanity check of our implementation, we confirmed that our average finetuning accuracy were comparable to prior work (\citet{sagawa2020} paper considers the task of predicting Blonde hair, with an ImageNet pretrained ResNet-50, they get 94.8\% and we get 95.9\% in this particular task, \ie when we also do the task of predicting Blonde hair with the same ImageNet pretrained ResNet-50). Further, on the \textbf{Wearing Earrings} task, we also confirmed that the CLIP pretrained ViT-B/16 did better than a ResNet-50 (both ImageNet pretrained and CLIP pretrained).
For each setting (\textit{scratch}, \textit{probing}, \textit{finetuning}), we trained 5 model runs with different seeds, for each of the tasks.
The final learning rates we used were 0.003 (training from scratch), 0.01 (linear probing), 0.000003 (fine-tuning), and they were selected by grid searching the learning rates on the \textbf{Earrings} task and we sanity checked that choosing a higher or lower learning rate led to lower accuracy.
We also train each approach for 10 epochs to ensure similar computational resources are provided to each approach.
In aggregate, all of the models we trained took approximately 1000 hours on one NVIDIA Titan Xp GPU. 

\subsubsection{Groupings}
\label{app:vision-groupings}

We consider groups based on hair color. 
The dataset provides five hair-related annotations of Black, Brown, Blonde, Grey, and Bald. 
Since the Bald category was quite small, we collapsed the category with all examples that lacked a hair annotation (\eg the hair color is obscured due to a hat) into an "Other" category to yield five total categories.
In addition, we measure outcome homogenization for individuals based on whether they have a \textit{beard}. 
While the \textbf{CelebA} dataset also contains annotations for race and gender, we chose to not look at these groups given we were concerned that the gender/race was being inferred by an annotator (crowdworker) from the face rather than being self-identified \citep{liu2015}.

\subsection{Language Experiments}
\label{app:language}

\subsubsection{Data}
\label{app:language-data}

We use the \textbf{IMDB}, \textbf{AGNews}, \textbf{Yahoo}, and \textbf{HateSpeech18} datasets.
For \textbf{IMDB}, we were unable to find formal license information.
The data may contain some PII, but it is unlikely there is significant offensive content. 
For \textbf{AGNews}, we were unable to find formal license information but found information indicating it should be used non-commercially, which we adhere to see.\footnote{See \url{http://groups.di.unipi.it/~gulli/AG_corpus_of_news_articles.html}.}
The data may contain some PII, but it is unlikely there is significant offensive content. 
For \textbf{Yahoo}, we were unable to find formal license information.
The data may contain some PII, especially given its nature, but it is unlikely there is significant offensive content. 
For \textbf{HateSpeech18}, we adhere to the license provided here: \url{https://github.com/Vicomtech/hate-speech-dataset#license}.
The data likely contains some PII given it is from forums, and certainly contains offensive content. 
We access this data through Hugging Face Datasets \citep{lhoest2021}.\footnote{\url{https://huggingface.co/datasets}}
The associated papers describe how the data was collected or scraped.
We tokenize the data using the RoBERTa \citep{liu2019} tokenizer provided in Hugging Face Transformers \citep{wolf2020}.

\subsubsection{Models}
\label{app:language-models}

For all models we produced, we adapt RoBERTa-base \citep{liu2019} using the weights provided through Hugging Face Transformers \citep{wolf2020}.
For all models, we use the default hyperparameters in the Trainer provide in the Transformer library, with the only change being a fairly standard setting of the learning rate to 2e-5. 
As a sanity check of our implementation, we confirmed that our accuracy matches those provided in standard scripts/tutorials provided in Transformers and are quite similar to other works that work with these standard datasets \citep[\eg][]{gururangan2019}.
For each setting (\textit{probing}, \textit{finetuning}, \textit{BitFit}), we trained 5 model runs with different seeds, for each of the four tasks.
In aggregate, all of the models we discuss in the paper took approximately 36 hours across 5 NVIDIA Titan Xp GPUs (or 180 hours on 1 NVIDIA Titan Xp GPU), with additional experiments/debugging that is unreported in the paper taking approximately an additional 2000 NVIDIA Titan Xp GPU hours.

\subsubsection{Groupings}
\label{app:language-groupings}
Since we consider four deployments that are largely unrelated to each other, there are no annotations of individuals or groups available that apply across all four datasets.
Consequently, we group inputs by (binary) gender, as this grouping applies across the four datasets.\footnote{We also considered grouping by \textit{names}, given recent works showing systemic behavior in NLP models for names \citep{shwartz2020, romanov2019}, and by \textit{race}, using names that are strongly statistically associated \citep{tzioumis2018, garg2018}. 
However, we found few systemic failures that we traced to the underlying groups: very few names appear in every dataset (often because the fictional movie characters and actors in \textbf{IDMB} are not discussed in the rest).}
Specifically, for each input we identify whether the input contains more references to the female gender (\eg uses of words like "she"), the male gender, or no reference to an explicitly gendered term is made.
We acknowledge that this treats gender as a binary as part of an unfortunate trend in NLP of works involving gender using binaries \citep{cao2020}.
We use the peer-reviewed list of gender terms from \citet{garg2018} and in accordance with the recommendations of \citet{antoniak2021}.
In the event that the same number of male and female gender terms are mentioned (possibly zero for both) in an input, we grouped the input in a third "Other" category.
While we did not extensively test, we did observe that the findings were not sensitive to small perturbations (\ie random deletions of words from each list) of the lists we used. \\
Following \citet{antoniak2021}, we provide the exact lists below. \\ \\
Male words = \{"he", "son", "his", "him", "father", "man", "boy", "himself", "male", "brother", "sons", "fathers", "men", "boys", "males", "brothers", "uncle", "uncles", "nephew", "nephews"\} \\
Female words = \{"she", "daughter", "hers", "her", "mother", "woman", "girl", "herself", "female", "sister", "daughters", "mothers", "women", "girls", "femen", "sisters", "aunt", "aunts", "niece", "nieces"\}

%% file: appendices/additional_experiments.tex
\section{Additional Experiments}
\label{app:experiments}


\paragraph{Summary.}
In \autoref{sec:data-sharing}, we report results on the \textbf{ACS PUMS} dataset.
Here, we supplement those findings to clarify whether the findings generalize across model families and to other datasets.
Qualitatively, across these additional evaluations, we do find the findings transfer: (i) the \textbf{fixed} partition of the data, where there is strictly greater data-sharing, reliably yields greater homogenization and (ii) when group-level data is available, individual-level homogenization exceeds group-level homogenization in magnitude.

\paragraph{Datasets.}
In \autoref{sec:data-sharing}, we report results on the \textbf{ACS PUMS} dataset.
Here, we replicate the experiments performed in that section, but vary the dataset to clarify if the qualitative trends generalize to other datasets.
Recall that the structure of the data we deal with is fairly unusual: we are interested in datasets where each input is associated with multiple outcomes (\ie the traditional multi-task learning setting) as we will share the training data across the models for each task.
However, because of our interests in social outcomes, we would further like each input to be meaningfully associated with a person (\eg arbitrary multi-task learning datasets could be used but are unideal if they don't additionally have this human-centric structure).

To identify additional relevant datasets, we survey datasets for multi-task learning, datasets for fairness in ML (which are generally human-centric as desired), and work at the intersection of fairness and multi-task ML.
Of these, we looked at \citet{zhang2017} for multi-task learning as well as \url{https://paperswithcode.com/task/multi-task-learning}, which provided a list of 51 datasets. 
We also looked at the 15 fairness datasets surveyed by \citet{quy2022}.
Finally, \citet{wang2021} initiated the study of multi-task fairness, considering four datasets.\footnote{Following our work, \citet{fabris2022} introduced a search engine for fairness datasets that confirms our selection process for datasets was comprehensive.}

From all of these, we arrived at four datasets with the structure we wanted: \textbf{CelebA}, \textbf{UCI Adult}, \textbf{LSAC} \citep{wightman1998}, and \textbf{GC} \citep[German Contracts;][]{dua2019}. 
All of these datasets have the desired structure: each input $x_j^i$ is directly associated with an individual $j$ and corresponds to multiple outcomes $y_j^1, \dots, y_j^k$.
Of them, we report results for \textbf{CelebA} in \autoref{subsec:fm-experiments}, and we use the \textbf{ACS PUMS} dataset of \citet{ding2021} that was explicitly designed to supersede the similar US Census-based \textbf{UCI Adult}.
Hence, we turn our attention to \textbf{GC} and \textbf{LSAC}.

The \textbf{GC} dataset contains information on 1000 German contracts,  including credit history, credit amount, and the corresponding credit risk for that individual \citep{dua2019}.
Following \citet{wang2021}, the two prediction tasks we consider are (i) predicting if the individual receives a good or bad loan and (ii) predicting whether their credit amount exceeds 2000.\footnote{We filter any individuals where only one of the outcomes is reported.}
We additionally featurize in the same way as \citet{wang2021}, using 16 attributes as features.
The data is accessed through the \texttt{UCI Machine Learning Repository}\footnote{https://archive.ics.uci.edu/ml/datasets/statlog+(german+credit+data)} and we use an 80/20 train-test split like \citet{wang2021}. 
All other experimental conditions match what we describe for the data-sharing experiments in \autoref{sec:data-sharing} with further details given in \autoref{app:census}.

The \textbf{LSAC} dataset was generated by the Law School Admission Council in the United States \citep{wightman1998}. 
This dataset contains information on 21,790 law students such as their entrance exam scores (LSAT) and their undergrad grade-point average (GPA) collected prior to law school.
From this, the two prediction tasks we consider are predicting (i) whether they pass the bar exam and (ii) whether their law school GPA exceeds the mean, directly following \citet{wang2021}.\footnote{We filter any students where only one of the outcomes is reported.}
The data is accessed through the \texttt{tempeh}\footnote{https://github.com/microsoft/tempeh} package with the default train-test split.
All other experimental conditions match what we describe for the data-sharing experiments in \autoref{sec:data-sharing} with further details given in \autoref{app:census}.

\paragraph{Model Families.}
In \autoref{sec:data-sharing}, we report results using logistic regression as the model family.
Here, we replicate the experiments performed in that section, but vary the model family to clarify the influence of the model family in the homogenization of outcomes.
Specifically, we consider three additional model families: gradient boosted decision tree classifiers (GBM),\footnote{Also considered by \citet{ding2021} in their work with \textbf{ACS PUMS}.} support vector machines, and neural networks. 
All of these models are implemented using \texttt{sk-learn} \citep{pedregosa2011} with default parameters.

\input{figures/census_partition_neurips_gbm.figure}
\input{figures/census_partition_neurips_svm.figure}
\input{figures/census_partition_neurips_nn.figure}

\input{figures/lsac_partition_neurips_logistic.figure}
\input{figures/lsac_partition_neurips_gbm.figure}
\input{figures/lsac_partition_neurips_svm.figure}

\input{figures/german_partition_neurips_logistic.figure}
\input{figures/german_partition_neurips_gbm.figure}
\input{figures/german_partition_neurips_svm.figure}

\paragraph{Results and Analysis.}
We report additional results for \textbf{ACS PUMS} in Figures~~\ref{fig:census-partition-gbm}--\ref{fig:census-partition-nn}, for \textbf{LSAC} in Figures~\ref{fig:lsac-partition-logistic}--\ref{fig:lsac-partition-svm}, and for \textbf{GC} in Figures~\ref{fig:german-partition-logistic}--\ref{fig:german-partition-nn}.
Across these figures, we first establish that our core findings are upheld: (i) homogenization is reliably greater when there is more homogenized (\textbf{fixed}) than less (\textbf{disjoint}) across datasets and model families and (ii) homogenization is reliably greater when contrasting individual-level measures with the appropriate racial group-level measures.\footnote{\textbf{ACS PUMS} is the only dataset where we have race metadata.}
However, we do note the absolute scale of homogenization is quite different across the datasets: we note that this is not surprising given the datasets are quite different, as are the relationship between the prediction tasks within a dataset.
Therefore, we highlight that our hypotheses make predictions about relative change (\ie sharing increases homogenization), but the underlying homogenization and absolute quantities will also depend significantly on the structure of the data and relationship between prediction tasks.

Further, for the \textbf{LSAC} and \textbf{GC} datasets, we visualize how the expected systemic failure rate (\ie the product of the error rates of the models, which is the denominator in our homogenization metric) changes as a function of data scale.
What we find is for both datasets, even one we have sufficient samples that the systemic failure rates in both the \textbf{disjoint} and \textbf{fixed} settings are identical (\ie variance due to finite sample effects in error rates becomes minimal), we see higher homogenization.
This demonstrates an important point: as the data grows, the expected systemic failure rate converges (as expected) to the same value for the \textbf{disjoint} and \textbf{fixed} partitions. 
That is, if one pays attention only to the accuracies for each task, these systems are the same.
But even at this state, we see sizable discrepancies in outcome homogenization (\ie the \textit{observed} systemic failure rates remain different) with the \textbf{fixed} partition displaying greater homogenization.
This drives home the point that data-sharing here has no effect on the accuracies of the resultant models, but that it does yield greater homogenization.
This demonstration is reminscent of work that demonstrates\footnote{Generally these works show the existence of such models is theoretically guaranteed. We encourage future work to provide similar guarantees for the systems we describe, beyond our initial empirical demonstrations} the existence of models of (near) equal accuracy but that are simpler \citep{semenova2022} or more fair \citep{marx2020}.
Here we observe a Rashomon effect \citep{breiman2001} at the level of social systems: both systems achieve the same accuracies, but one (\textbf{disjoint}) is more homogeneous than the other (\textbf{fixed}). 

%% file: figures/census_partition_neurips_gbm.figure.tex
\begin{figure}[t]
\centering
\includegraphics[width=\linewidth]{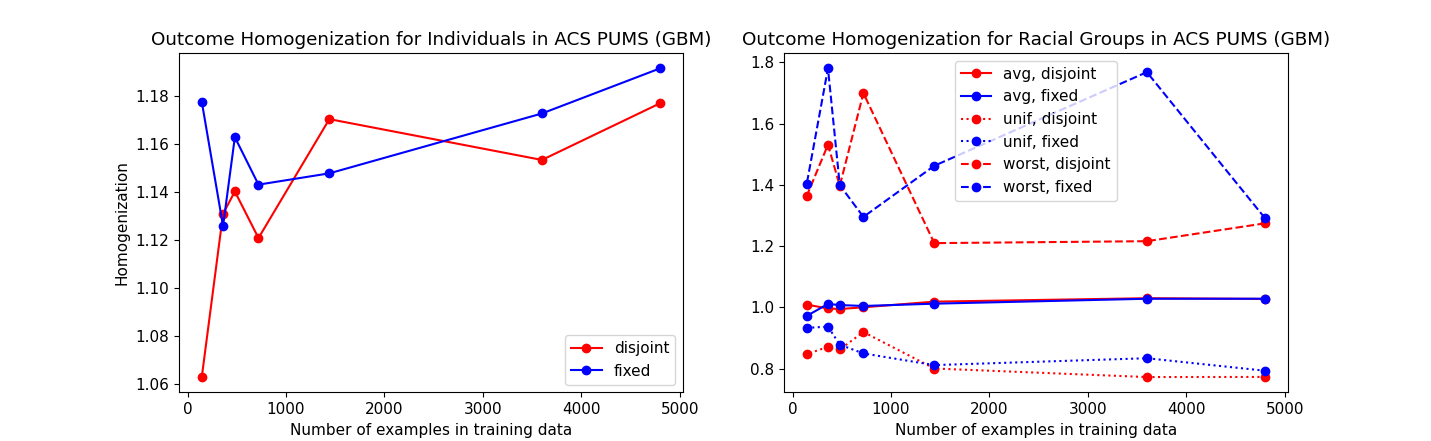}
\caption{
Results for data-sharing experiments on \textbf{ACS PUMS} with gradient boosted classifiers showing homogenization ($y$) as a function of training dataset size ($x$). 
Training across tasks on the same data (\textbf{fixed}) yields more homogeneous outcomes than on non-identical but identically distributed data (\textbf{disjoint}), especially for small datasets.
}
\label{fig:census-partition-gbm}
\end{figure}

%% file: figures/census_partition_neurips_svm.figure.tex
\begin{figure}[t]
\centering
\includegraphics[width=\linewidth]{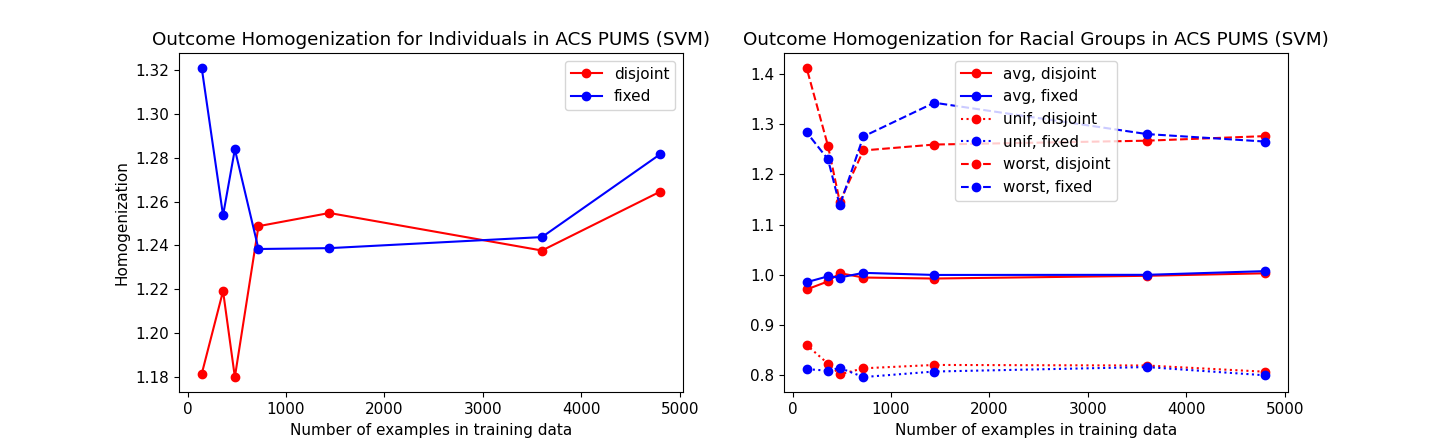}
\caption{
Results for data-sharing experiments on \textbf{ACS PUMS} with support vector machines showing homogenization ($y$) as a function of training dataset size ($x$). 
Training across tasks on the same data (\textbf{fixed}) yields more homogeneous outcomes than on non-identical but identically distributed data (\textbf{disjoint}), especially for small datasets.
}
\label{fig:census-partition-svm}
\end{figure}

%% file: figures/census_partition_neurips_nn.figure.tex
\begin{figure}[t]
\centering
\includegraphics[width=\linewidth]{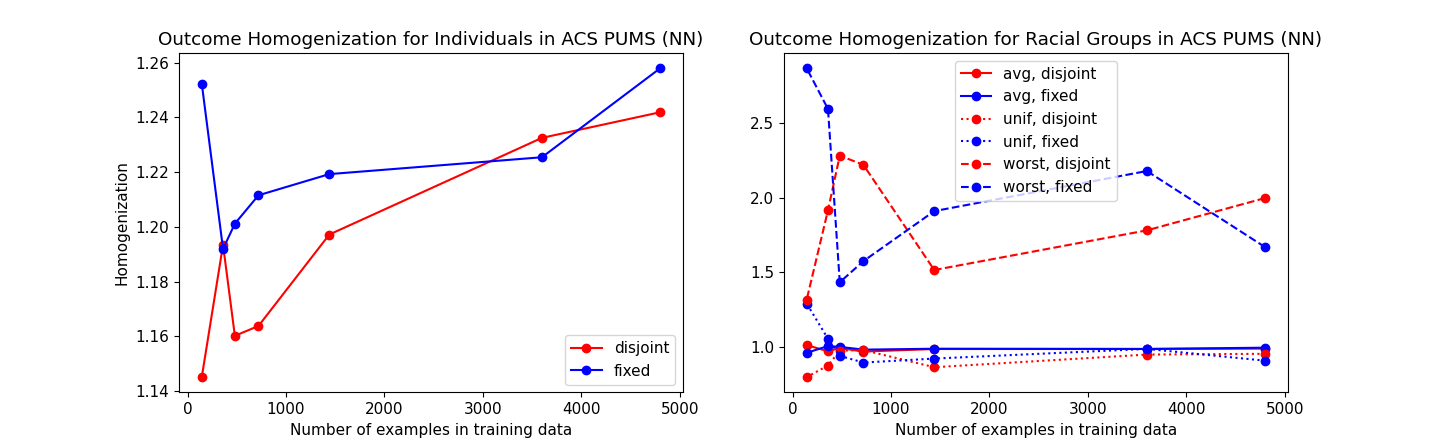}
\caption{
Results for data-sharing experiments on \textbf{ACS PUMS} with neural network classifiers showing homogenization ($y$) as a function of training dataset size ($x$). 
Training across tasks on the same data (\textbf{fixed}) yields more homogeneous outcomes than on non-identical but identically distributed data (\textbf{disjoint}), especially for small datasets.
}
\label{fig:census-partition-nn}
\end{figure}

%% file: figures/lsac_partition_neurips_logistic.figure.tex
\begin{figure}[t]
\centering
\includegraphics[width=\linewidth]{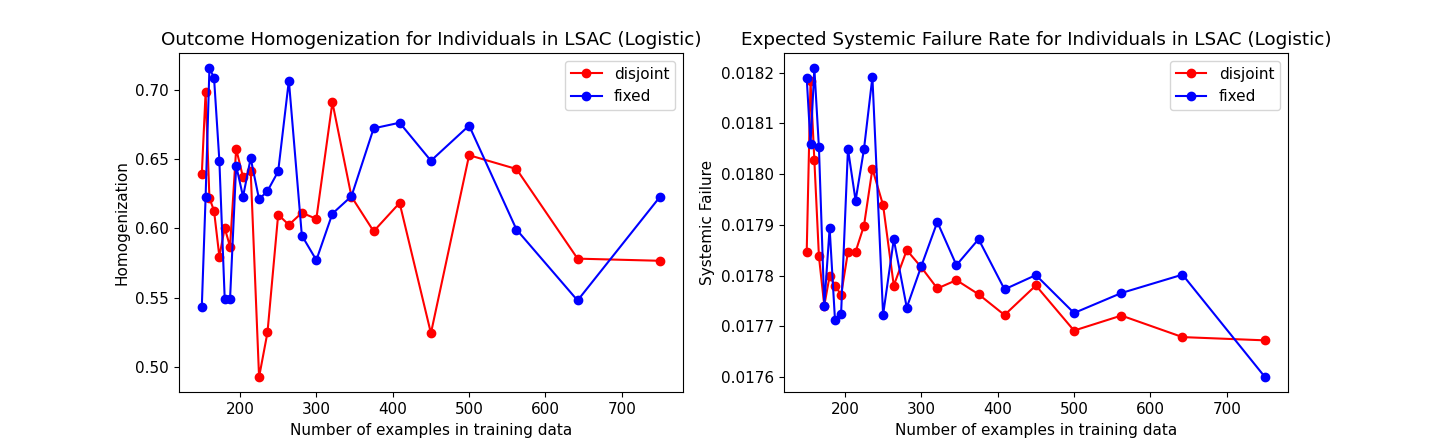}
\caption{
Results for data-sharing experiments on \textbf{LSAC} with logistic regression classifiers showing homogenization (\textbf{left}) and expected systemic failure rate $(\prod\limits_{i \in [k]}~\fail(h^i); \textbf{ right})$, which is the denominator in homogenization,  as a function of training dataset size ($x$). 
}
\label{fig:lsac-partition-logistic}
\end{figure}

%% file: figures/lsac_partition_neurips_gbm.figure.tex
\begin{figure}[t]
\centering
\includegraphics[width=\linewidth]{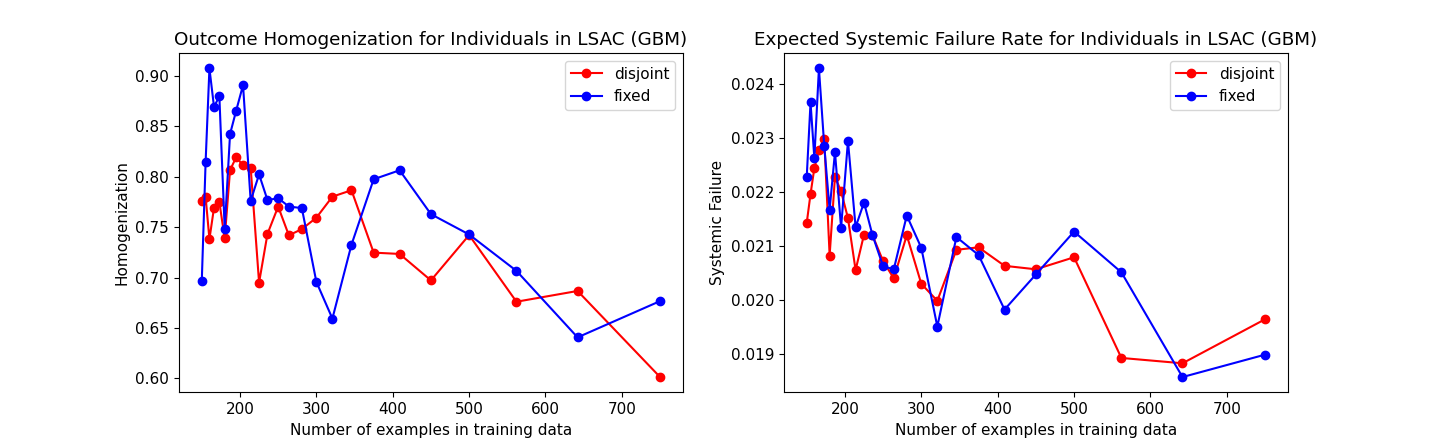}
\caption{
Results for data-sharing experiments on \textbf{LSAC} with gradient boosted classifiers showing homogenization (\textbf{left}) and expected systemic failure rate $(\prod\limits_{i \in [k]}~\fail(h^i); \textbf{ right})$, which is the denominator in homogenization,  as a function of training dataset size ($x$). 
}
\label{fig:lsac-partition-gbm}
\end{figure}

%% file: figures/lsac_partition_neurips_svm.figure.tex
\begin{figure}[t]
\centering
\includegraphics[width=\linewidth]{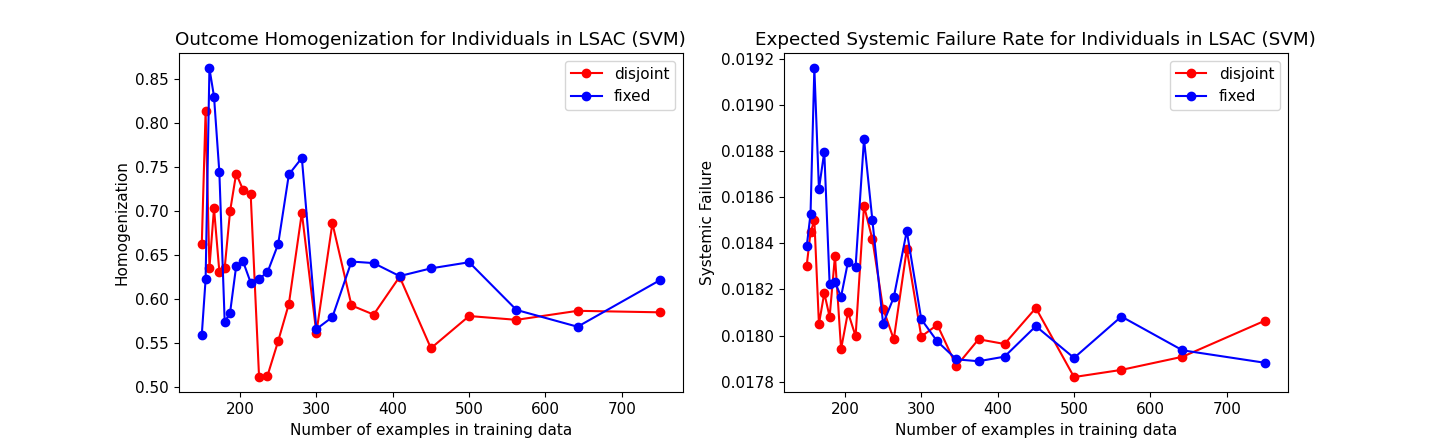}
\caption{
Results for data-sharing experiments on \textbf{LSAC} with support vector machines showing homogenization (\textbf{left}) and expected systemic failure rate $(\prod\limits_{i \in [k]}~\fail(h^i); \textbf{ right})$, which is the denominator in homogenization,  as a function of training dataset size ($x$). 
}
\label{fig:lsac-partition-svm}
\end{figure}

%% file: figures/german_partition_neurips_logistic.figure.tex
\begin{figure}[t]
\centering
\includegraphics[width=\linewidth]{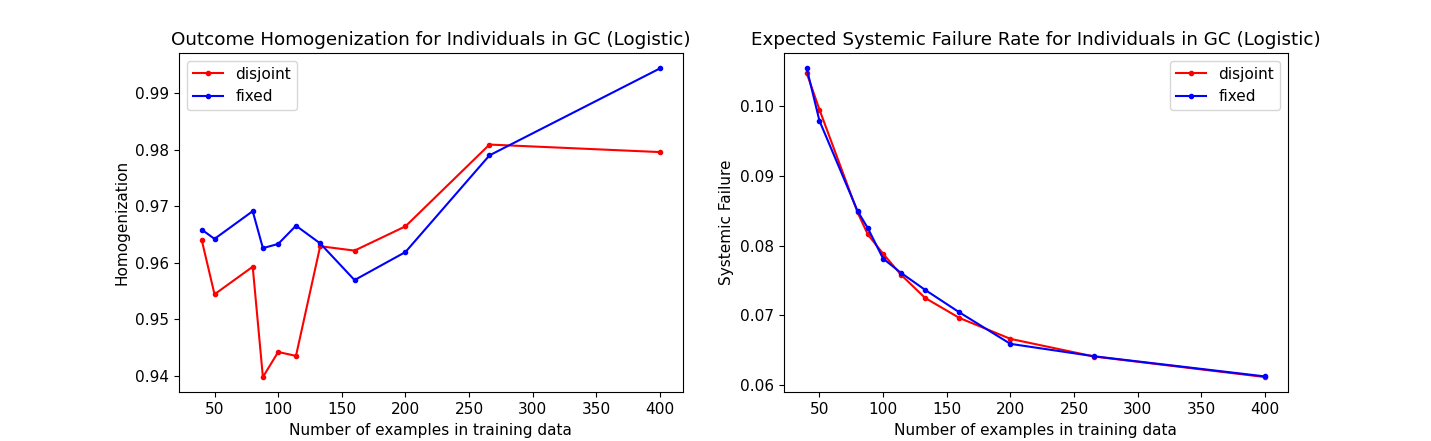}
\caption{
Results for data-sharing experiments on \textbf{GC} with logistic regression showing homogenization (\textbf{left}) and expected systemic failure rate $(\prod\limits_{i \in [k]}~\fail(h^i); \textbf{ right})$, which is the denominator in homogenization,  as a function of training dataset size ($x$). 
}
\label{fig:german-partition-logistic}
\end{figure}

%% file: figures/german_partition_neurips_gbm.figure.tex
\begin{figure}[t]
\centering
\includegraphics[width=\linewidth]{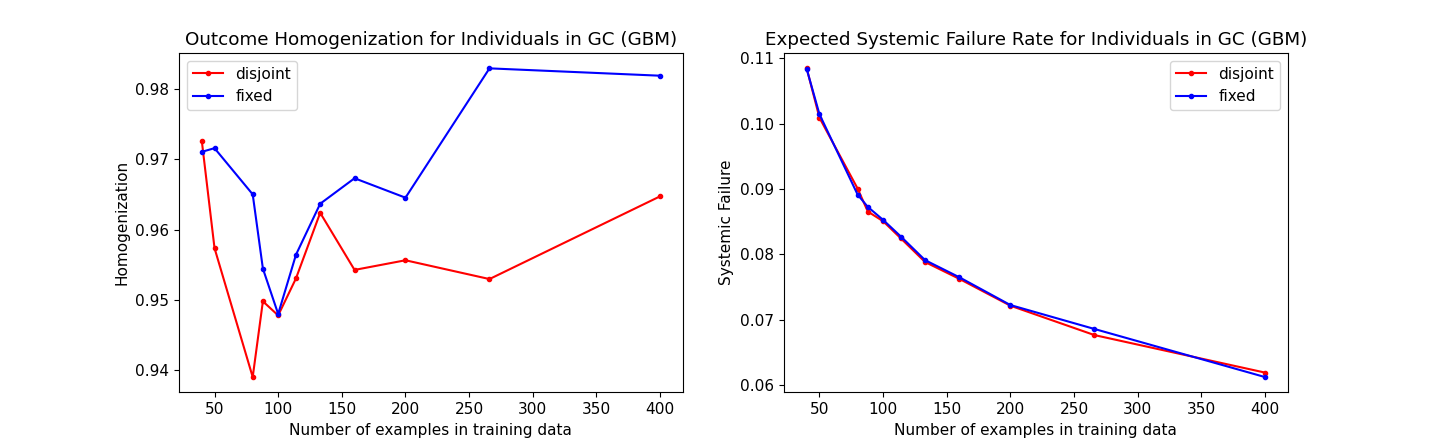}
\caption{
Results for data-sharing experiments on \textbf{GC} with gradient boosted classifiers showing homogenization (\textbf{left}) and expected systemic failure rate $(\prod\limits_{i \in [k]}~\fail(h^i); \textbf{ right})$, which is the denominator in homogenization,  as a function of training dataset size ($x$). 
}
\label{fig:german-partition-gbm}
\end{figure}

%% file: figures/german_partition_neurips_svm.figure.tex
\begin{figure}[t]
\centering
\includegraphics[width=\linewidth]{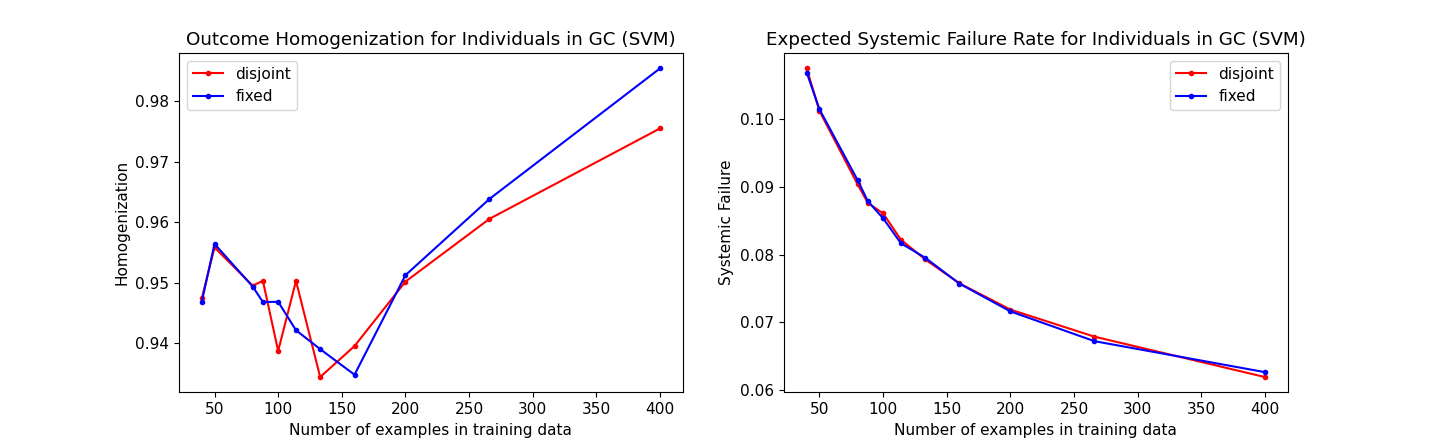}
\caption{
Results for data-sharing experiments on \textbf{GC} with support vector machines showing homogenization (\textbf{left}) and expected systemic failure rate $(\prod\limits_{i \in [k]}~\fail(h^i); \textbf{ right})$, which is the denominator in homogenization,  as a function of training dataset size ($x$). 
}
\label{fig:german-partition-svm}
\end{figure}